\documentclass{article}
\PassOptionsToPackage{table}{xcolor}

\usepackage{iclr2027_conference,times} 

\usepackage{amsmath,amsfonts,bm}

\def\eqref#1{equation~\ref{#1}}

\def\1{\bm{1}}

\DeclareMathAlphabet{\mathsfit}{\encodingdefault}{\sfdefault}{m}{sl}
\SetMathAlphabet{\mathsfit}{bold}{\encodingdefault}{\sfdefault}{bx}{n}

\usepackage{hyperref}
\usepackage{url}
\usepackage{graphicx}
\usepackage{booktabs}
\usepackage{caption}
\usepackage{subcaption}

\iclrfinalcopy 

\usepackage{tabularx}
\usepackage{enumitem}
\usepackage[most]{tcolorbox}
\usepackage{xspace}

\usepackage{xcolor}
\usepackage[normalem]{ulem} 

\newcommand{\method}{{\em WASP-Mem}\xspace}

\newcommand{\cond}[1]{\uline{#1}}

\graphicspath{{figures/}}

\definecolor{OurRow}{HTML}{FFF0E8}
\definecolor{ModelRow}{HTML}{EEF2F6}

\newtcolorbox{promptwrapper}[1]{
  enhanced,
  breakable,
  colback=gray!5,
  colframe=black!60,
  boxrule=0.6pt,
  arc=3pt,
  left=5pt,
  right=5pt,
  top=3pt,
  bottom=3pt,
  title=#1,
  coltitle=white,
  colbacktitle=black!75,
  fonttitle=\bfseries,
  attach boxed title to top left={yshift=-2mm, xshift=3mm},
  boxed title style={
    boxrule=0pt,
    arc=3pt,
    outer arc=3pt
  }
}

\title{\method: Let Agent Organize Procedural Memory. Human Spend Effort on Diagnosis}

\author{%
Peizhi Niu$^{1}$\thanks{Equal contribution.} \quad
Ge Shi$^{2}$\footnotemark[1] \quad
Vaibhav Gorde$^{2}$ \quad
Sarthak Ghosh$^{2}$ \quad
Christos Faloutsos$^{2}$ \\
Min Cheng$^{2}$ \quad
Kevin Chen$^{2}$ \\[4pt]
$^{1}$University of Illinois Urbana-Champaign \quad $^{2}$Amazon \\
\texttt{peizhin2@illinois.edu}, \\
\texttt{\{gegshi, gordev, ghossart, faloutso, mmcheng, kevnchen\}@amazon.com}
}

\begin{document}

\maketitle
\lhead{Preprint} 

\begin{abstract}
  LLM agents struggle to automate complex enterprise procedures because static, pre-built harness cannot capture proprietary policies, undocumented behaviors, or state-dependent recovery paths that surface only during execution. Procedural memory accrued from agent experience offers a training-free path to closing this gap—yet unlike episodic records or semantic facts, it must consolidate repeated successes and diagnosed failures into reusable actions and recovery strategies. We introduce \method (\(\mathcal{W}\)iki-based \(\mathcal{A}\)gent-\(\mathcal{S}\)ynthesized \(\mathcal{P}\)rocedural \(\mathcal{M}\)emory), a nimble but effective method that distills execution traces and feedback into persistent, interlinked wiki pages, enabling continual agent learning. On TheAgentCompany benchmark, WASP-Mem raises the checkpoint pass rate by \textbf{4.6–6.1}\% over the strongest baseline we compared across three models while requiring \textbf{57–61}\% less storage. We further present a systematic controlled study spanning consolidation method, feedback type, trace ingestion scope, and refinement rounds. Our results reveal that prescriptive knowledge organization such as human defined schemas and explicit deduplication rules hurts performance rather than closing the procedural-knowledge gap. In contrast, human diagnosis converts failed traces into corrective knowledge and drives sustained improvement: a single round of human diagnosis outperforms four rounds of self-judgment by \textbf{7.0–10.9}\%. These findings point to a simple division of labor: let agents organize procedural memory, and invest human effort where it matters most—diagnosis.
\end{abstract}

\section{Introduction}
\label{sec:introduction}

LLM agents are becoming practical interfaces for complex enterprise procedure
automation: they plan, call tools, and modify shared services to complete
multi-step workplace tasks \citep{theagentcompany,selfevolving}. Their behavior
is governed by a static agent harness, including the software implementation,
system prompt, tools, environment interfaces, and fixed contexts supplied
before deployment \citep{coala}. Yet the operational knowledge needed in a
proprietary environment is intrinsically unforeseeable: under-documented APIs,
organization-specific policies, strict output conventions, abnormal service
states, and recovery paths often surface only during execution. In production,
designers cannot know in advance which hidden condition will cause the next
failure. Compounding this, unlike many coding or mathematical benchmarks where
broadly available knowledge combines with compilers, tests, exact answers, or
resettable sandboxes to provide cheap verification
\citep{swebench,mathdataset}, enterprise feedback depends on dedicated
proprietary environments and organization-specific outcomes, making evaluation
and diagnosis substantially more expensive \citep{theagentcompany}.

Long-term memory offers a training-free way to close this residual knowledge
gap without changing model parameters \citep{coala,memsurvey}. Episodic memory
preserves what happened in a particular rollout; semantic memory stores general
facts or concepts. Procedural memory, in contrast, captures how to act under
particular conditions
\citep{tulving1972,cohen1980,squire1992,anderson1993}. Its unit is not a
transcript or isolated statement but reusable guidance specifying when a
procedure applies, which actions and implementation details carry it out, what
can fail, and how to recover. Consequently, procedural memory requires
consolidation rather than naive accumulation: repeated successes should be
merged, contradictions resolved, and diagnosed failures incorporated as
missing preconditions or troubleshooting paths
\citep{automanual,awm,memp}. Without such consolidation, raw sessions accumulate
redundancy and noise without becoming accurate, robust knowledge that
generalizes across recurring task instances.

\begin{figure}[tb]
\centering
\begin{subfigure}[b]{0.56\linewidth}
  \centering
  \includegraphics[height=4.5cm,width=\linewidth,keepaspectratio]{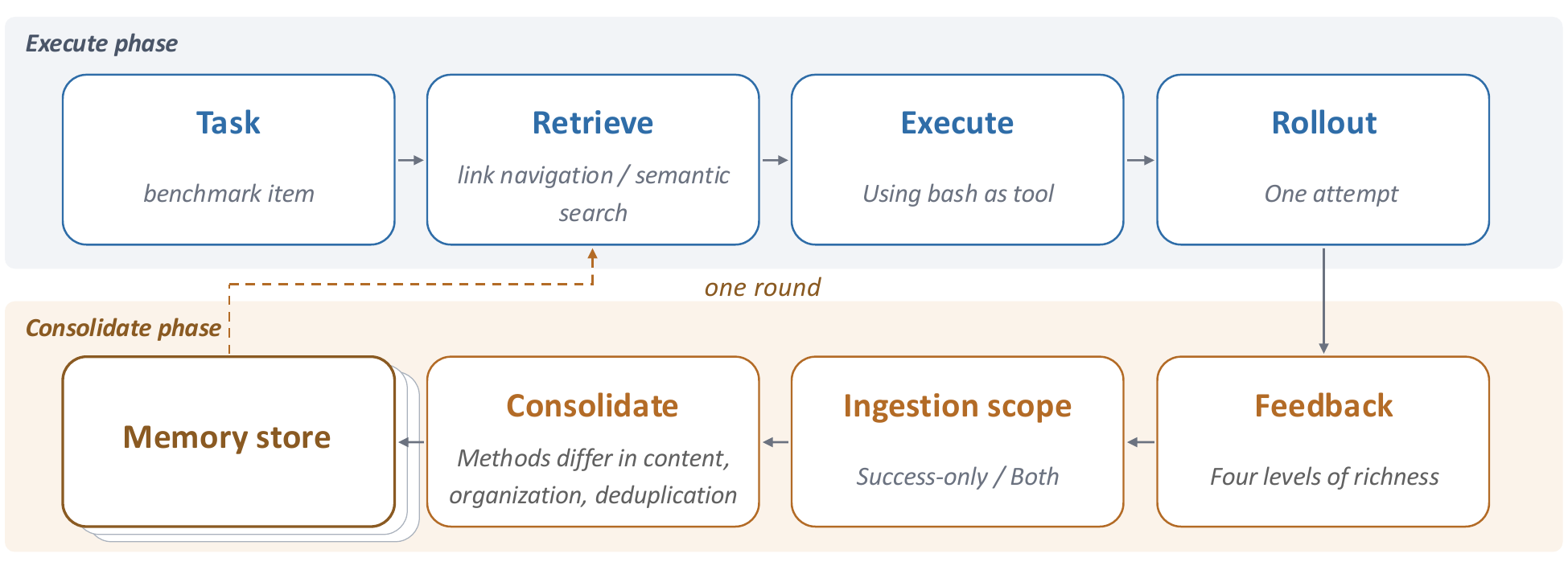}
  \caption{Procedural memory consolidation workflow.}
  \label{fig:overview-workflow}
\end{subfigure}%
\hfill%
\begin{subfigure}[b]{0.43\linewidth}
  \centering
  \includegraphics[height=4.5cm,width=\linewidth,keepaspectratio]{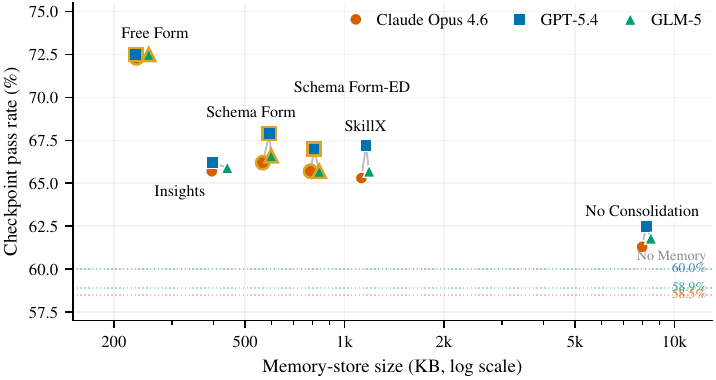}
  \caption{Memory-store size vs.\ checkpoint pass rate.}
  \label{fig:filter-size}
\end{subfigure}
\caption{\textbf{Workflow and key results.}
(a)~The multi-round loop enables continual learning by consolidate agent execution traces and feedback into reusable procedural memory.
(b)~Memory-store size and checkpoint performance across consolidation methods
and model variants. \method achieves higher checkpoint pass rate with smaller memory store size.}
\label{fig:combined}
\end{figure}

Formally, this process is a multi-round execution--consolidation cycle (Figure~\ref{fig:overview-workflow}). During
the execution phase, given a task, the agent retrieves zero or more relevant
memory pages, acts on the environment through its tools, and records its actions
and observations as a raw rollout trace. An evaluator or human then provides
feedback based on the resulting environment state. During the consolidation
phase, a method reads the current store alongside eligible traces and feedback,
and abstracts them into reusable procedural knowledge: when a procedure
applies, which actions to perform, what failures may arise, and how to diagnose
and resolve them. The method then writes or merges this knowledge according to
the memory system's representation and interaction interface. One execution
pass followed by consolidation forms a round; only the updated memory store
carries into the next round, while the model parameters and harness remain
fixed.

Agent memory is a rapidly expanding design space, with new representations,
extraction rules, retrieval mechanisms, and update procedures appearing
continuously
\citep{memsurvey,selfevolving,amem,reasoningbank,insights,skillx}. Choosing
among overwhelming candidates resembles a classic Secretary Problem
\citep{ferguson1989secretary} with no known stopping rule, and exhaustive
evaluation is infeasible. Practitioners instead need higher-level principles
explaining which design dimensions matter, where human and system effort
produces the greatest return, and which interventions merely add constraints or
volume. We organize our study around three questions: \emph{(1) How should execution experience be consolidated into reusable
procedural memory? (2) Which feedback, trace ingestion scope, and refinement
choices most improve future execution? (3) Where should human effort be spent:
prescribing knowledge organization upfront or diagnosing failures reactively?}

Through controlled experiments and qualitative analysis, we derive actionable
answers to these questions. Our contributions are threefold:

\begin{itemize}[leftmargin=*,itemsep=2pt,topsep=3pt]
    \item \textit{Nimble but Effective Consolidation Method.} We introduce
    \(\mathcal{W}\)iki-based \(\mathcal{A}\)gent-\(\mathcal{S}\)ynthesized \(\mathcal{P}\)rocedural \(\mathcal{M}\)emory (\method), an agentic
    procedural-memory organizer for training-free continual learning. It
    consolidates agent execution traces and human feedback into persistent,
    interlinked wiki pages, achieving both the most compact knowledge store and
    the strongest downstream task performance among all compared methods
    (Figure~\ref{fig:filter-size}).
    \item \textit{Controlled Study.} To determine what makes procedural-memory
    consolidation effective, we systematically vary consolidation method,
    feedback type, trace-ingestion scope, and refinement rounds across three
    models, identifying which interventions most improve future execution.
    \item \textit{Learnings From a Bitter Lesson.} We distill our findings into
    a design principle: build a general-purpose memory
    system, but let the agent self-organize its procedural knowledge. Imposing
    domain-specific structure on procedural knowledge at design time harms
    performance. Instead, human intervention is most effective when applied
    reactively---through failure diagnosis, targeted feedback, and iterative
    refinement.
\end{itemize}

\section{Related Work}
\label{sec:related}

\paragraph{Procedural memory for continual learning.}
Many long-term agent-memory systems focus on episodic or semantic memories and are evaluated through conversational or question-answering recall
\citep{memgpt,mem0,amem,coala,memsurvey}, rather than reusable procedural
knowledge. Procedural approaches broadly follow three patterns. Case-based
methods retain and retrieve trajectories or task-local reflections
\citep{reflexion,memento}; these preserve exact evidence but are
case-specific, sensitive data leakage risky and accumulate redundancy without cross-session consolidation. Abstractive methods distill traces into reusable
insights \citep{expel,autoguide,insights} but processes rollouts independently, without cross-session deduplication or verification. Skill-based methods induce workflows, scripts, or hierarchical routines \citep{voyager,awm,memp,skillx}, yet reusable components can be duplicated across skills and conflicts may go unresolved. In contrast, for \method, agent synthesizes interlinked procedural pages at different granularities, and contextually deduplicates and reconciles knowledge. The resulting memory is compact navigable, and readable by both agents and humans.

\paragraph{Benchmarks for enterprise procedural learning.}
Suitable benchmarks are scarce because they must expose proprietary-knowledge
gaps, target enterprise automation rather than coding or mathematics, provide
stateful tools, include enough related tasks for procedural transfer, and offer execution annotations finer than task-level success.
TheAgentCompany provides multi-service professional workflows and
checkpoint-level grading \citep{theagentcompany}, while $\tau^2$-Bench  provides domain policies, tool APIs, verifiable state transitions, and customer-support tasks \citep{tau2bench}.
SaaS-Bench broadens the setting to larger scale but its strongest reported agent completes fewer
than 4\% of tasks end-to-end \citep{saasbench} and package was also unavailable during our experiments. We adopt TheAgentCompany and $\tau^2$-Bench: agents generate rollouts and feedback on training tasks, each method consolidates them into memory, and the agent with frozen store is evaluated on held-out test tasks. We judge memory methods by downstream
checkpoint and task completion, rather than recall of stored experience.

\section{Method}
\label{sec:method}

We target recurring enterprise workflows in which a task attempt can
modify a shared environment and relevant organization-specific procedures
may be absent from the initial agent harness.
Let $M_r$ denote the persistent memory store at round $r$, $B_r$ a batch
of execution traces, $F_r$ the corresponding environment or human
feedback, and $S$ the ingestion scope governing which traces and feedback
are eligible for consolidation. A consolidation method $C$ updates the
store as
\begin{equation}
    M_{r+1}=C(M_r,B_r,F_r;S).
    \label{eq:consolidation}
\end{equation}
The goal is to capture residual procedural knowledge not already encoded
in the fixed harness. Consolidation transforms many rollout-specific
observations into fewer reusable procedures by merging repeated evidence,
correcting inaccurate knowledge, and retaining distinct lessons when their
conditions or remedies differ. The model parameters, harness, and executor
loop remain fixed; methods differ only in their memory representation and
consolidation operation. After consolidation on training tasks, the store
is frozen and evaluated by the checkpoint and task completion rates the
agent achieves on test tasks.

\subsection{\method{} Consolidation Method}
\label{sec:wasp-method}

\method{} is a memory-consolidation and retrieval method for training-free
continual learning that organizes execution traces and feedback into
persistent, interlinked procedural wiki pages at different granularities and
retrieves them to guide future execution, drawing on the idea of an
LLM-maintained wiki as a personal knowledge base \citep{karpathy2026llmwiki}.
\method{} initializes a wiki filesystem with no initial procedural content and
mandatory control files such as \texttt{index.md}, which serves as a
high-level router over the wiki pages. A \method{} consolidator agent creates,
organizes, revises, or removes pages and their links during consolidation.
Three interface operations are provided:

\begin{itemize}[leftmargin=*,itemsep=2pt,topsep=3pt]
    \item \textbf{Ingestion.} Given a raw agent-session \texttt{.jsonl} file
    and optional asynchronously produced natural-language feedback, the
    \method{} agent spawns a subagent that reviews the session trace to
    identify actionable observations retrospectively, then uses them to add,
    update, or remove pages and links.
    \item \textbf{Query.} During both consolidation and execution, agents begin
    from \texttt{index.md} and combine semantic search with shell navigation to
    locate relevant pages and follow inter-page relationships, such as related
    conditions, actions, and outcomes; the semantic-search index is rebuilt
    after each consolidation phase.
    \item \textbf{Lint.} Prompt-based rules and scripts direct the agent to
    scrutinize generalizability, reusability, ambiguity, excessive page length,
    link integrity, and conflicts, and to determine whether pages should be
    split, merged, or relinked.
\end{itemize}

\paragraph{\method{} variants.}
\emph{Free Form} is the basic \method{} instantiation described above. Each
page is markdown with a minimal metadata header, such as its title, brief
description, and last-modified time; the consolidator decides the remaining
page structure and content. \emph{Schema Form} adds prescriptive,
human-designed organization: pages must be Procedures, Actions, or Lessons,
each with fixed fields and enumerated options. Both \emph{Free Form} and
\emph{Schema Form} read the existing wiki before writing and therefore perform
implicit, agent-managed deduplication. \emph{Schema Form-ED} further adds explicit deduplication
rules: each session is first processed independently into candidate pages,
after which a separate normalization, matching, and merge stage consolidates
candidates before committing them to the persistent store. \emph{Schema Form}
and \emph{Schema Form-ED} thus represent increasing levels of upfront
prescriptive knowledge organization.

\subsection{Controlled Consolidation Design Space}
\label{sec:design-space}

To identify which consolidation choices improve downstream execution and where
human and system effort is most rewarding, we vary four dimensions
systematically while holding the remaining factors fixed.

\paragraph{Consolidation method.}
We compare the three \method{} variants above with four baselines. \emph{No
Memory} executes without assistance from any long-term memory, while \emph{No
Consolidation} stores raw execution traces and retrieves the most similar trace
without abstraction or integration. \emph{Insights} distills each session into
abstract procedural insights \citep{insights} and stores them in a flat
collection, with no cross-session deduplication, verification, or
reconciliation. \emph{SkillX} constructs a knowledge base in the form of agent
skills organized at three levels: planning, functional, and atomic
\citep{skillx}. This hierarchy is prescriptive, analogous to imposing a page
schema: each planning skill forms an independent procedural hierarchy, and
reusable functional or atomic skills are not deduplicated into a shared pool
across those hierarchies.

\paragraph{Feedback type.}
Feedback determines what evidence the consolidator has for deciding whether an
execution was correct and why it succeeded or failed. We construct a ladder
from \emph{Self-Judge}, which uses only the agent's own assessment, to external
\emph{Checkpoint-binary} feedback that localizes success and failure to
individual checkpoints within the task, and \emph{Checkpoint-binary + Human}
feedback that additionally provides natural-language diagnosis. This separates
the value of an external outcome signal from the value of diagnostic
explanation.

\paragraph{Ingestion scope.}
Ingestion scope determines which experiences are eligible for consolidation.
\emph{Success-only} admits evidence from successful executions, whereas
\emph{Both} also admits failed traces and their feedback. Failure traces become
useful when outcome or diagnostic feedback distinguishes a reusable procedural
lesson from incidental unsuccessful behavior.

\paragraph{Refinement rounds.}
Refinement rounds control whether memory is consolidated once or repeatedly
revised as further execution traces become available. R0 serves as the
no-memory baseline, R1 performs the first consolidation, and R2--R4 repeatedly
execute against and update the same persistent store, allowing knowledge to be
corroborated, corrected, or extended over time. Appendix~\ref{app:methods}
specifies the complete controlled matrix and protocols.

\section{Experiments}
\label{sec:experiments}

\subsection{Experimental Setup}
\label{sec:setup}

\paragraph{TheAgentCompany.}
Our primary controlled study uses TheAgentCompany
\citep{theagentcompany}, which contains 127 workplace tasks and 542
execution-based checkpoints across GitLab, RocketChat, OwnCloud, and Plane.
Its six professional domains share tools, APIs, and workflows, providing
recurring procedural structure across tasks. The isolated services form a
resettable testbed for state-changing workplace actions. To model recurring
enterprise automation in which agents repeatedly encounter the same workflows,
training and test sets contain the same 127 tasks. We enforced strict post-consolidation guardrails to prevent case-specific data leakage. Each condition scores one
fresh attempt per task under the current memory store.

\paragraph{$\tau^2$-Bench Retail.}
We additionally use the retail domain of $\tau^2$-Bench
\citep{tau2bench} to examine in-domain generalization across disjoint train and
test sets. We define a custom operating procedure that is absent from the agent
harness and encode its requirements in the ground-truth action checkpoints. The
disjoint train and test tasks share substantial procedural knowledge through
common policy rules and tool workflows. The resulting benchmark contains 74
training tasks with 391 checkpoints and 40 test tasks with 192 checkpoints. Agents first execute the training
tasks without access to the procedure. The resulting traces and checkpoint
feedback are then used to construct each memory condition's store, which is
frozen and evaluated on the test tasks while the original procedure remains
withheld. 

\paragraph{Models, harnesses, and environments.}
For both benchmarks, we report results with Claude Opus 4.6 \citep{claudeopus46}, GPT-5.4 \citep{gpt54}, and
GLM-5 \citep{glm5}. On TheAgentCompany, we use the \texttt{pi-agent}
harness, which exposes bash as its only action tool for issuing API calls
against self-hosted services in isolated Docker environments. Cohere Embed v4
provides 1024-dimensional embeddings for semantic search. For $\tau^2$-Bench,
we use its native function-calling agent and simulated customer environment.
All memory methods receive the same training rollout batch, and LLM-based
methods use the same consolidator model
(Appendix~\ref{app:reproducibility}).

\paragraph{Metrics and controls.}
Our primary metric is \emph{checkpoint pass rate}: the fraction of binary,
execution-grounded checkpoints passed. For $\tau^2$-Bench, checkpoints are
ground-truth action sequences verified against the environment state.
\emph{Fully-passed rate} is the fraction of tasks satisfying all benchmark
evaluation criteria. We also report memory-store size and consolidation API
cost. The full four-dimensional controlled study is conducted on
TheAgentCompany, while $\tau^2$-Bench repeats the consolidation-method
comparison under held-out in-domain generalization.
Appendix~\ref{app:results} gives the complete matrix and exact results, and
Appendix~\ref{app:reproducibility} reports implementation details and costs.

\subsection{Consolidation Method}
\label{sec:exp1}

\begin{table}[tb]
\caption{\textbf{Consolidation method across model variants.}
Bold marks the best value for each metric within a block, and the shaded
row is the Free Form \method.}
\label{tab:exp1}
\centering
\small
\resizebox{\textwidth}{!}{
\begin{tabular}{l ccc ccc}
\toprule
& \multicolumn{3}{c}{\textsc{TheAgentCompany}} & \multicolumn{3}{c}{$\tau^2$\textsc{-Bench}} \\
\cmidrule(lr){2-4} \cmidrule(lr){5-7}
Method & Checkpoint pass & Fully passed & Store (KB) & Checkpoint pass & Fully passed & Store (KB) \\
\midrule
\rowcolor{ModelRow}
\multicolumn{7}{l}{\textbf{Claude Opus 4.6}} \\
No Memory          & 58.5\%                          & 36.2\%                          & n/a     & 66.7\%                          & 40.0\%                          & n/a     \\
No Consolidation   & 61.3\% {\scriptsize(+2.8)}      & 40.2\% {\scriptsize(+4.0)}      & 7{,}979 & 69.3\% {\scriptsize(+2.6)}      & 42.5\% {\scriptsize(+2.5)}      & 1{,}428 \\
SkillX             & 65.3\% {\scriptsize(+6.8)}      & 45.7\% {\scriptsize(+9.5)}      & 1{,}125 & \textbf{81.2\%} {\scriptsize(+14.5)} & 50.0\% {\scriptsize(+10.0)}     & 336     \\
Insights           & 65.7\% {\scriptsize(+7.2)}      & 45.7\% {\scriptsize(+9.5)}      & 396     & 78.1\% {\scriptsize(+11.4)}     & \textbf{55.0\%} {\scriptsize(+15.0)} & 96      \\
\cmidrule(lr){1-7}
\multicolumn{7}{l}{\textit{WASP-Mem}} \\
\rowcolor{OurRow}
\quad Free Form (Implicit Dedup)             & \textbf{72.3\%} {\scriptsize(+13.8)} & 48.0\% {\scriptsize(+11.8)}     & \textbf{234} & \textbf{81.2\%} {\scriptsize(+14.5)} & \textbf{55.0\%} {\scriptsize(+15.0)} & \textbf{92}  \\
\quad + Schema (Implicit Dedup)              & 66.2\% {\scriptsize(+7.7)}     & \textbf{48.8\%} {\scriptsize(+12.6)} & 565          & 77.6\% {\scriptsize(+10.9)}     & 42.5\% {\scriptsize(+2.5)}      & 252          \\
\quad + Schema + Explicit Dedup              & 65.7\% {\scriptsize(+7.2)}     & 44.9\% {\scriptsize(+8.7)}      & 789          & 79.7\% {\scriptsize(+13.0)}     & 52.5\% {\scriptsize(+12.5)}     & 656          \\
\midrule
\rowcolor{ModelRow}
\multicolumn{7}{l}{\textbf{GPT-5.4}} \\
No Memory          & 60.0\%                          & 40.2\%                          & n/a     & 65.1\%                          & 35.0\%                          & n/a     \\
No Consolidation   & 62.5\% {\scriptsize(+2.5)}      & 42.5\% {\scriptsize(+2.3)}      & 8{,}208 & 62.0\% {\scriptsize($-$3.1)}    & 42.5\% {\scriptsize(+7.5)}      & 1{,}953 \\
SkillX             & 67.2\% {\scriptsize(+7.2)}      & 50.4\% {\scriptsize(+10.2)}     & 1{,}161 & 69.3\% {\scriptsize(+4.2)}      & 45.0\% {\scriptsize(+10.0)}     & 425     \\
Insights           & 66.2\% {\scriptsize(+6.2)}      & 48.8\% {\scriptsize(+8.6)}      & 399     & 67.7\% {\scriptsize(+2.6)}      & 52.5\% {\scriptsize(+17.5)}     & \textbf{112} \\
\cmidrule(lr){1-7}
\multicolumn{7}{l}{\textit{WASP-Mem}} \\
\rowcolor{OurRow}
\quad Free Form (Implicit Dedup)             & \textbf{72.5\%} {\scriptsize(+12.5)} & \textbf{52.0\%} {\scriptsize(+11.8)} & \textbf{233} & \textbf{76.6\%} {\scriptsize(+11.5)} & \textbf{60.0\%} {\scriptsize(+25.0)} & 123          \\
\quad + Schema (Implicit Dedup)              & 67.9\% {\scriptsize(+7.9)}     & 50.4\% {\scriptsize(+10.2)}     & 592          & 71.4\% {\scriptsize(+6.3)}      & 47.5\% {\scriptsize(+12.5)}     & 303          \\
\quad + Schema + Explicit Dedup              & 67.0\% {\scriptsize(+7.0)}     & 49.6\% {\scriptsize(+9.4)}      & 808          & 71.9\% {\scriptsize(+6.8)}      & 52.5\% {\scriptsize(+17.5)}     & 880          \\
\midrule
\rowcolor{ModelRow}
\multicolumn{7}{l}{\textbf{GLM-5}} \\
No Memory          & 58.9\%                          & 38.6\%                          & n/a     & 47.4\%                          & 25.0\%                          & n/a     \\
No Consolidation   & 61.8\% {\scriptsize(+2.9)}      & 40.9\% {\scriptsize(+2.3)}      & 8{,}480 & 46.4\% {\scriptsize($-$1.0)}    & 30.0\% {\scriptsize(+5.0)}      & 1{,}461 \\
SkillX             & 65.7\% {\scriptsize(+6.8)}      & 46.5\% {\scriptsize(+7.9)}      & 1{,}187 & 53.1\% {\scriptsize(+5.7)}      & 32.5\% {\scriptsize(+7.5)}      & 315     \\
Insights           & 65.9\% {\scriptsize(+7.0)}      & 47.2\% {\scriptsize(+8.6)}      & 441     & 51.6\% {\scriptsize(+4.2)}      & 37.5\% {\scriptsize(+12.5)}     & \textbf{82}  \\
\cmidrule(lr){1-7}
\multicolumn{7}{l}{\textit{WASP-Mem}} \\
\rowcolor{OurRow}
\quad Free Form (Implicit Dedup)             & \textbf{72.5\%} {\scriptsize(+13.6)} & \textbf{52.0\%} {\scriptsize(+13.4)} & \textbf{255} & \textbf{58.3\%} {\scriptsize(+10.9)} & \textbf{42.5\%} {\scriptsize(+17.5)} & 92           \\
\quad + Schema (Implicit Dedup)              & 66.6\% {\scriptsize(+7.7)}     & 50.4\% {\scriptsize(+11.8)}     & 599          & 56.2\% {\scriptsize(+8.8)}      & 37.5\% {\scriptsize(+12.5)}     & 323          \\
\quad + Schema + Explicit Dedup              & 65.7\% {\scriptsize(+6.8)}     & 46.5\% {\scriptsize(+7.9)}      & 838          & 53.1\% {\scriptsize(+5.7)}      & 27.5\% {\scriptsize(+2.5)}      & 557          \\
\bottomrule
\end{tabular}}
\end{table}

\subsubsection{Main Results}
\label{sec:exp1-main-results}

This experiment tests whether consolidating experience improves future execution
and how the organization of procedural memory affects transfer. We compare No
Memory, raw-trace retrieval without consolidation, Insights, SkillX, and the
three \method{} variants. All conditions use checkpoint-binary feedback,
ingest both successful and failed traces, and consolidate in a single round
(R1). Within each benchmark, all memory methods receive the same training
rollout batch and are evaluated under the same execution budget.

Table~\ref{tab:exp1} shows that consolidation, rather than experience storage
alone, drives improvement. Every consolidated method exceeds No Memory across
all six model--benchmark combinations. In contrast, No Consolidation provides
only +2.5--2.9 pp on TheAgentCompany and changes checkpoint pass rate by
$-3.1$ to +2.6 pp on $\tau^2$-Bench. Raw traces can therefore preserve useful
examples, but without abstraction and reconciliation they transfer
inconsistently and can even distract execution.

Free Form \method{} provides the strongest and most consistent result. It leads
checkpoint pass rate for every model on TheAgentCompany and leads or ties the
best baseline on $\tau^2$-Bench (tying SkillX for Claude at 81.2\%), improving
by +10.9--14.5 pp over No Memory across the six settings. It also improves
fully-passed rate by +11.8--25.0 pp and achieves the best or tied-best value
in five of six settings. These gains do not come from retaining more
experience: its 92--255 KB stores are approximately 94--97\% smaller than No
Consolidation. Free Form is the smallest consolidated store for all three
TheAgentCompany models and the smallest \method{} variant on $\tau^2$-Bench.

The \method{} ablations further show that prescriptive knowledge organization
does not improve consolidation. Adding a human-defined schema lowers checkpoint
pass rate relative to Free Form in every model--benchmark combination by
2.1--6.1 pp, while increasing store size by 2.3--3.5$\times$. Adding explicit
deduplication also remains 1.5--6.8 pp below Free Form and expands the store by
3.3--7.2$\times$. Thus, agent-managed implicit deduplication produces a more
compact and effective procedural memory than either prescribed page types or
prescribed cross-session matching rules.

Figure~\ref{fig:filter-size} reinforces that a compact memory store with high signal density outperforms a large one
filled with redundancy and noise. \emph{These results
suggest that memory consolidation acts as a filter. Its value lies not in
volume but in selectivity, retaining informative procedural signal while
discarding what is false, noisy, redundant, or misleading.} Under this view,
we identify two failure modes:

\begin{itemize}[leftmargin=*,itemsep=2pt,topsep=3pt]
    \item \textbf{Under-filtering} retains noise, redundancy, and incidental
    detail. No Consolidation stores entire rollouts without abstraction.
    Insights and SkillX abstract individual sessions but process them
    independently, so similar procedures with nuanced differences produce
    separate entries rather than being merged into a shared, deduplicated
    procedure.
    \item \textbf{Over-filtering} discards useful signal. Schema Form's fixed
    fields can exclude implementation details not anticipated when the schema
    was designed. SkillX's rigid three-level hierarchy may similarly drop
    contextual information that does not fit its prescribed structure.
\end{itemize}

\noindent
These two modes can co-occur: a method may simultaneously retain redundant
entries (under-filtering) while discarding actionable detail that falls outside
its prescribed structure (over-filtering), producing a mis-filtered store.
Free Form \method{} achieves a more balanced filter: reading the existing wiki
before writing enables implicit deduplication, while unconstrained pages
preserve whichever actionable details execution reveals, yielding the strongest
and most consistent checkpoint pass rate across model variants.

\begin{promptwrapper}{Let Agents Organize Memory.}
\textbf{Prescriptive knowledge organization does not improve consolidation. Agent management produces more compact and effective procedural memory.}
\end{promptwrapper}

\subsection{Feedback Type and Ingestion Scope}
\label{sec:exp23}

This experiment asks how feedback richness and trace-ingestion scope jointly
affect consolidation quality. It uses Free Form \method{} consolidated in a
single round (R1). Feedback varies from \emph{Self-Judge}, to external
\emph{Checkpoint-binary} outcomes, to \emph{Checkpoint-binary + Human}
feedback with natural-language diagnosis. Within each feedback condition, we
compare \emph{Success-only} ingestion with \emph{Both}, which also admits
failed traces. The two scopes consolidate the same R0 rollout batch and
receive the same subsequent execution budget, so their difference isolates the
value of learning from failure.

\begin{figure}[tb]
\centering
\begin{subfigure}[t]{0.43\linewidth}
    \centering
    \includegraphics[width=\linewidth]{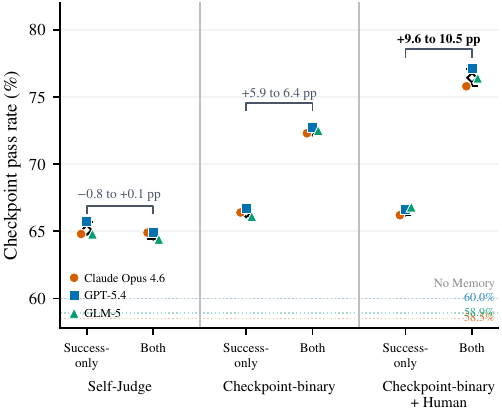}
    \caption{Checkpoint pass rate.}
    \label{fig:ingestion-scope-checkpoint}
\end{subfigure}\hfill
\begin{subfigure}[t]{0.43\linewidth}
    \centering
    \includegraphics[width=\linewidth]{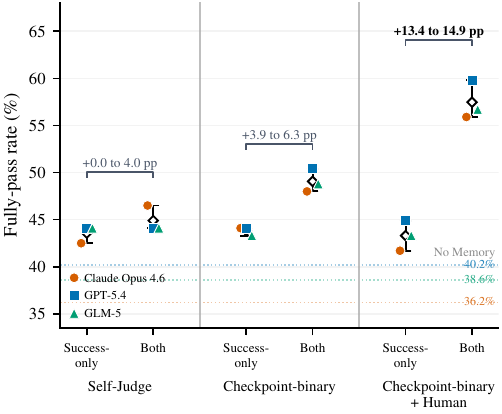}
    \caption{Full pass rate.}
    \label{fig:ingestion-scope-full}
\end{subfigure}
\caption{\textbf{Feedback type and ingestion scope.}
Checkpoint pass rate (a) and fully-passed rate (b) when consolidating
successful traces only versus both successful and failed traces under
Self-Judge, Checkpoint-binary, and Checkpoint-binary + Human feedback.
Colored points denote model variants; black diamonds and whiskers show the
cross-model mean and min--max range.}
\label{fig:feedback-scope}
\end{figure}

Feedback richness first determines the overall quality of the consolidated
memory. When both successful and failed traces are ingested, checkpoint pass
rate rises from 64.4--64.9\% with Self-Judge to 72.3--72.7\% with
Checkpoint-binary and 75.8--77.1\% with Checkpoint-binary + Human. External
checkpoint outcomes improve performance by +7.4--8.1 pp over Self-Judge by
correcting mistaken self-assessments and localizing unsuccessful actions.
Human diagnosis adds a further +3.5--4.4 pp by explaining failure causes and
corrective procedures that cannot be recovered reliably from a binary outcome
alone.

This benefit is concentrated in failed traces. Under Success-only ingestion,
moving from Self-Judge to Checkpoint-binary + Human changes checkpoint pass
rate by only +0.9--2.0 pp, and adding human diagnosis beyond Checkpoint-binary
changes it by $-0.2$ to +0.7 pp. Under Both, the same comparisons yield
+10.9--12.2 pp and +3.5--4.4 pp, respectively. Fully-passed rate exhibits the
same interaction: rich feedback changes Success-only by only $-0.8$ to
+0.8 pp over Self-Judge, but improves Both by +9.4--15.7 pp. Diagnostic
feedback therefore contributes primarily by making failed executions
interpretable, rather than by adding substantial information to traces that
already succeeded.

Viewed from the other axis, richer feedback determines how much the agent can
learn from failed traces (Figure~\ref{fig:feedback-scope}). Moving from
Success-only to Both changes checkpoint pass rate by only $-0.7$ to +0.1 pp
under Self-Judge. Without reliable external signal, the agent cannot
distinguish a reusable lesson from incidental unsuccessful behavior, so
admitting failures adds noise rather than knowledge. The gain grows to
+5.9--6.4 pp with Checkpoint-binary feedback and +9.6--10.5 pp with
Checkpoint-binary + Human feedback. The fully-passed rate exhibits the same
progression: admitting failed traces adds 0.0--4.0 pp under Self-Judge,
+3.9--6.3 pp under Checkpoint-binary, and +13.4--14.9 pp under
Checkpoint-binary + Human. Failure traces are therefore not automatically
useful: they become valuable when feedback identifies what failed and, most
effectively, why it failed and how the procedure should be corrected.

\begin{promptwrapper}{Feedback Makes Failure Learnable.}
\textbf{The richer and more diagnostic the feedback, the more downstream value
the agent extracts from failed traces.}
\end{promptwrapper}

\subsection{Multi-Round Refinement}
\label{sec:multi-round}

This experiment asks whether repeated execution and consolidation continue to
close the residual procedural-knowledge gap. It uses Free Form \method{} with
Both ingestion scope, varies rounds from R0 through R4, and compares
Self-Judge, Checkpoint-binary, and Checkpoint-binary + Human feedback. R0
executes without memory; each subsequent round executes against the current
frozen store and then consolidates the resulting traces and feedback. All
reported results reflect fresh executions from that round rather than
cumulative or best-over-round scores.

\begin{figure}[tb]
\centering
\begin{subfigure}[t]{0.43\linewidth}
    \centering
    \includegraphics[width=\linewidth]{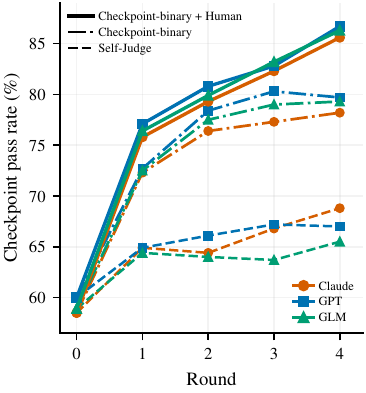}
    \caption{Reactive refinement.}
    \label{fig:rounds-refinement}
\end{subfigure}\hfill
\begin{subfigure}[t]{0.5\linewidth}
    \centering
    \includegraphics[width=\linewidth]{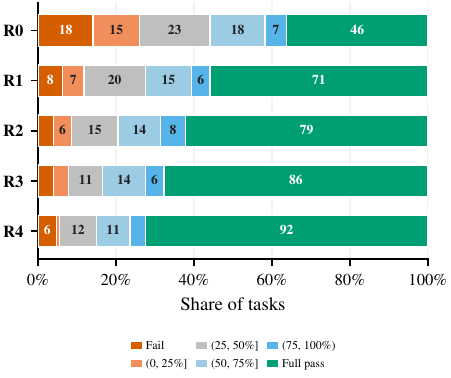}
    \caption{Progress toward full pass.}
    \label{fig:rounds-completion}
\end{subfigure}
\caption{\textbf{Multi-round refinement and task completion.}
(a) Checkpoint pass rate across R0--R4 for all three models under
Checkpoint-binary + Human, Checkpoint-binary, and Self-Judge feedback.
(b) Claude Opus 4.6 task scores under Checkpoint-binary + Human feedback.
Segment labels are task counts.}
\label{fig:rounds}
\end{figure}

With Checkpoint-binary + Human feedback, checkpoint pass rate increases
monotonically for every model and reaches 85.6--86.7\% at R4. The first
consolidation round contributes +17.1--17.5 pp, yet the residual failures
continue to reveal useful procedural gaps: R3 to R4 still adds +3.1--3.9 pp,
with gains remaining non-trivial through the final round. Checkpoint-binary
feedback reaches only 78.2--79.7\% at R4, with its final-round change
shrinking to $-0.6$ to +0.9 pp. Self-Judge is lower and non-monotonic, ending
at 65.5--68.8\%. Human diagnosis therefore does more than improve the first
memory store; it repeatedly turns the remaining failures into new corrective
knowledge.

Feedback quality also matters more than additional rounds alone. After one
Checkpoint-binary + Human round, checkpoint pass rate is already
75.8--77.1\%, exceeding the corresponding Self-Judge R4 result by
7.0--10.9 pp for every model. Four rounds of self-assessment do not recover
the information supplied by one round of external outcomes and human failure
diagnosis.

Figure~\ref{fig:rounds-completion} shows how local checkpoint gains translate
into end-to-end task success. For Claude Opus 4.6, fully passed tasks increase
from 46 at R0 to 71, 79, 86, and 92 through R4. As repeated refinement makes
earlier steps reliable and resolves later failure points, tasks shift from
partial completion into full passes.

\begin{promptwrapper}{Feedback Quality over Rollout Volume.}
\textbf{Richer feedback on fewer traces outperforms sparse feedback on more
rollouts; local checkpoint gains compound into end-to-end success.}
\end{promptwrapper}

\section{Discussion}
\label{sec:discussion}

Our study answers the three questions posed in the introduction and turns them
into a design principle for agentic continual learning.

\paragraph{How should experience be consolidated?}
Experience should be transformed into shared, revisable procedural memory
rather than accumulated as independent traces or abstractions. By organizing
interlinked pages, \method{} can merge and correct knowledge across sessions
while preserving useful relationships at different granularities. Free Form
\method{} improves over No Memory by 10.9--14.5 pp across all six
model--benchmark settings with 92--255 KB stores---two orders of magnitude
smaller than raw-trace retrieval. The value of memory lies in selective
consolidation, not volume.

\paragraph{Which feedback, ingestion, and refinement choices matter?}
Failure traces are useful primarily when feedback makes them interpretable.
Execution-grounded outcomes localize failures, diagnosis explains the required
correction, and iterative refinement turns newly exposed gaps into durable
knowledge. One diagnostically supervised round outperforms four Self-Judge
rounds by 7.0--10.9 pp: feedback quality matters more than rollout volume.

\paragraph{Where should human effort be spent?}
General system design remains necessary: memory needs persistence, retrieval,
update operations, navigation, and quality checks. Domain-specific
organization, however, should be left to the agent rather than prescribed
before the knowledge gap is observed. Human effort is better spent reactively
diagnosing failures. Additional constraints should address demonstrated memory
failures, not anticipated ones.

\method{} therefore contributes both a training-free mechanism for acquiring
proprietary procedural knowledge and a practical principle: let the agent
organize memory, and spend human effort on diagnosis.

\paragraph{Limitations.}
Our controlled study uses two benchmarks, three models, and single-run
configurations; variance across seeds remains unquantified
(Appendix~\ref{app:limitations}). More broadly, enterprise procedural
automation remains less studied than code, mathematics, or personal assistants,
and no available open benchmark combines stateful environments, proprietary
procedures, and execution-grounded feedback at scale. Longer-term memory
stability, broader enterprise environments, replication, and scalable automated
diagnosis remain important directions for future work.

\section{Conclusion}
\label{sec:conclusion}

We introduced \method{}, a training-free continual-learning method that
consolidates execution traces and feedback into persistent, interlinked
procedural wiki pages. Across two enterprise benchmarks and three model
families, Free Form \method{} leads or ties checkpoint pass rate in all six
settings while maintaining compact stores. The controlled study further shows
that prescribed schemas and explicit deduplication rules do not improve
consolidation, that failed traces become useful as feedback becomes more
diagnostic, and that one round with checkpoint outcomes and human diagnosis
outperforms four rounds of self-judgment. Continued diagnosis produces larger
and more sustainable improvements across refinement rounds, with local
checkpoint corrections compounding into end-to-end task completion. The
resulting design principle is simple: provide a general memory system, let the
agent organize the procedural knowledge revealed by experience, and spend human
effort reactively on diagnosing the failures that remain.

\newpage

\section*{AI Use Statement}
Generative AI is both the object of study in this work and a tool we used to
carry it out. As the object of study, large language models execute the
benchmark tasks, perform consolidation, and supply \cond{Self-Judge} feedback
in the corresponding condition, as described in Section~\ref{sec:method} and
Appendix~\ref{app:methods}. As a tool, we used an LLM-based assistant to give
feedback on experimental design, help interpret results, implement analysis and
plotting scripts, draft and edit the manuscript, and identify and format
related work. We did not use it to generate data or produce any reported number
outside the experimental runs themselves. All AI-generated or AI-assisted
outputs were reviewed and verified by the authors, who take full responsibility
for the paper's content, claims, results, and artifacts.

\section*{Ethics Statement}
We identify no ethical concerns with this work. It involves no human-subjects
research and releases no personal data. Every experiment ran against
self-hosted instances of GitLab, RocketChat, OwnCloud, and Plane populated
with the benchmark's synthetic content; no production system or real user data
was accessed. The \cond{Checkpoint-binary + Human} condition required a
qualified annotator to describe each failed checkpoint; that annotator produced
labels for the study rather than being a subject of it.

\section*{Reproducibility Statement}
Section~\ref{sec:method} defines the consolidation cycle and the four
dimensions we vary; Section~\ref{sec:setup} states the benchmarks, models,
harness, metrics, and controls held fixed. Appendix~\ref{app:methods} provides
per-method consolidation recipes, representation-specific retrieval adapters,
feedback definitions, and the complete experimental matrix.
Appendix~\ref{app:results} reports every per-model result behind the ranges
quoted in the main text. Appendix~\ref{app:reproducibility} records the
environment, model versions, retrieval configuration, compute, wall-clock
time, per-batch API cost, round procedure, and experiment code. Every
configuration was run once; Appendix~\ref{app:limitations} states which
differences we therefore treat as unresolved.

\clearpage

\bibliography{references}

\begin{thebibliography}{31}
\providecommand{\natexlab}[1]{#1}
\providecommand{\url}[1]{\texttt{#1}}
\expandafter\ifx\csname urlstyle\endcsname\relax
  \providecommand{\doi}[1]{doi: #1}\else
  \providecommand{\doi}{doi: \begingroup \urlstyle{rm}\Url}\fi

\bibitem[Anderson(1993)]{anderson1993}
John~R. Anderson.
\newblock \emph{Rules of the Mind}.
\newblock Lawrence Erlbaum Associates, 1993.

\bibitem[ang Gao et~al.(2025)ang Gao, Geng, Hua, Hu, Juan, Liu, Liu, Qiu, Qi,
  Wu, Wang, Xiao, Zhou, Zhang, Zhang, Xiang, Fang, Zhao, Liu, Ren, Qian, Wang,
  Hu, Wang, Wu, Ji, and Wang]{selfevolving}
Huan ang Gao, Jiayi Geng, Wenyue Hua, Mengkang Hu, Xinzhe Juan, Hongzhang Liu,
  Shilong Liu, Jiahao Qiu, Xuan Qi, Yiran Wu, Hongru Wang, Han Xiao, Yuhang
  Zhou, Shaokun Zhang, Jiayi Zhang, Jinyu Xiang, Yixiong Fang, Qiwen Zhao,
  Dongrui Liu, Qihan Ren, Cheng Qian, Zhenhailong Wang, Minda Hu, Huazheng
  Wang, Qingyun Wu, Heng Ji, and Mengdi Wang.
\newblock A survey of self-evolving agents: What, when, how, and where to
  evolve on the path to artificial super intelligence.
\newblock \emph{arXiv preprint arXiv:2507.21046}, 2025.
\newblock \doi{10.48550/arXiv.2507.21046}.
\newblock URL \url{https://arxiv.org/abs/2507.21046}.

\bibitem[{Anthropic}(2026)]{claudeopus46}
{Anthropic}.
\newblock Claude opus 4.6 system card.
\newblock \url{https://www.anthropic.com/claude-opus-4-6-system-card}, February
  2026.

\bibitem[Barres et~al.(2025)Barres, Dong, Ray, Si, and Narasimhan]{tau2bench}
Victor Barres, Honghua Dong, Soham Ray, Xujie Si, and Karthik Narasimhan.
\newblock {$\tau^2$-Bench}: Evaluating conversational agents in a dual-control
  environment.
\newblock \emph{arXiv preprint arXiv:2506.07982}, 2025.
\newblock \doi{10.48550/arXiv.2506.07982}.
\newblock URL \url{https://arxiv.org/abs/2506.07982}.

\bibitem[Chen et~al.(2024)Chen, Li, Yang, Yu, Lin, and He]{automanual}
Minghao Chen, Yihang Li, Yanting Yang, Shiyu Yu, Binbin Lin, and Xiaofei He.
\newblock Automanual: Constructing instruction manuals by llm agents via
  interactive environmental learning.
\newblock \emph{arXiv preprint arXiv:2405.16247}, 2024.
\newblock \doi{10.48550/arXiv.2405.16247}.
\newblock URL \url{https://arxiv.org/abs/2405.16247}.

\bibitem[Chhikara et~al.(2025)Chhikara, Khant, Aryan, Singh, and Yadav]{mem0}
Prateek Chhikara, Dev Khant, Saket Aryan, Taranjeet Singh, and Deshraj Yadav.
\newblock {Mem0}: Building production-ready {AI} agents with scalable long-term
  memory.
\newblock \emph{arXiv preprint arXiv:2504.19413}, 2025.
\newblock \doi{10.48550/arXiv.2504.19413}.
\newblock URL \url{https://arxiv.org/abs/2504.19413}.

\bibitem[Cohen \& Squire(1980)Cohen and Squire]{cohen1980}
Neal~J. Cohen and Larry~R. Squire.
\newblock Preserved learning and retention of pattern-analyzing skill in
  amnesia: Dissociation of knowing how and knowing that.
\newblock \emph{Science}, 210\penalty0 (4466):\penalty0 207--210, 1980.

\bibitem[Du et~al.(2025)Du, Huang, Zheng, Wang, Montella, Lapata, Wong, and
  Pan]{memsurvey}
Yiming Du, Wenyu Huang, Danna Zheng, Zhaowei Wang, Sebastien Montella, Mirella
  Lapata, Kam-Fai Wong, and Jeff~Z. Pan.
\newblock Rethinking memory in llm based agents: Representations, operations,
  and emerging topics.
\newblock \emph{arXiv preprint arXiv:2505.00675}, 2025.
\newblock \doi{10.48550/arXiv.2505.00675}.
\newblock URL \url{https://arxiv.org/abs/2505.00675}.

\bibitem[Fang et~al.(2025)Fang, Liang, Wang, Wu, Qiao, Xie, Huang, Chen, and
  Zhang]{memp}
Runnan Fang, Yuan Liang, Xiaobin Wang, Jialong Wu, Shuofei Qiao, Pengjun Xie,
  Fei Huang, Huajun Chen, and Ningyu Zhang.
\newblock Memp: Exploring agent procedural memory.
\newblock \emph{arXiv preprint arXiv:2508.06433}, 2025.
\newblock \doi{10.48550/arXiv.2508.06433}.
\newblock URL \url{https://arxiv.org/abs/2508.06433}.
\newblock Published at ACL 2026 Findings.

\bibitem[Ferguson(1989)]{ferguson1989secretary}
Thomas~S. Ferguson.
\newblock Who solved the secretary problem?
\newblock \emph{Statistical Science}, 4\penalty0 (3):\penalty0 282--289, 1989.
\newblock \doi{10.1214/ss/1177012493}.

\bibitem[Fu et~al.(2024)Fu, Kim, Kim, Sohn, Logeswaran, Bae, and
  Lee]{autoguide}
Yao Fu, Dong-Ki Kim, Jaekyeom Kim, Sungryull Sohn, Lajanugen Logeswaran,
  Kyunghoon Bae, and Honglak Lee.
\newblock Autoguide: Automated generation and selection of context-aware
  guidelines for large language model agents.
\newblock \emph{arXiv preprint arXiv:2403.08978}, 2024.
\newblock \doi{10.48550/arXiv.2403.08978}.
\newblock URL \url{https://arxiv.org/abs/2403.08978}.
\newblock Published at NeurIPS 2024.

\bibitem[{GLM-5 Team}(2026)]{glm5}
{GLM-5 Team}.
\newblock {GLM-5}: from vibe coding to agentic engineering.
\newblock \emph{arXiv preprint arXiv:2602.15763}, 2026.
\newblock URL \url{https://arxiv.org/abs/2602.15763}.

\bibitem[Hendrycks et~al.(2021)Hendrycks, Burns, Kadavath, Arora, Basart, Tang,
  Song, and Steinhardt]{mathdataset}
Dan Hendrycks, Collin Burns, Saurav Kadavath, Akul Arora, Steven Basart, Eric
  Tang, Dawn Song, and Jacob Steinhardt.
\newblock Measuring mathematical problem solving with the {MATH} dataset.
\newblock \emph{arXiv preprint arXiv:2103.03874}, 2021.
\newblock \doi{10.48550/arXiv.2103.03874}.
\newblock URL \url{https://arxiv.org/abs/2103.03874}.
\newblock Published at NeurIPS 2021 Datasets and Benchmarks Track.

\bibitem[Hu et~al.(2026)Hu, Long, and Wang]{insights}
Qisheng Hu, Quanyu Long, and Wenya Wang.
\newblock When continual learning moves to memory: A study of experience reuse
  in llm agents.
\newblock \emph{arXiv preprint arXiv:2604.27003}, 2026.
\newblock \doi{10.48550/arXiv.2604.27003}.
\newblock URL \url{https://arxiv.org/abs/2604.27003}.

\bibitem[Jimenez et~al.(2023)Jimenez, Yang, Wettig, Yao, Pei, Press, and
  Narasimhan]{swebench}
Carlos~E. Jimenez, John Yang, Alexander Wettig, Shunyu Yao, Kexin Pei, Ofir
  Press, and Karthik Narasimhan.
\newblock {SWE-bench}: Can language models resolve real-world {GitHub} issues?
\newblock \emph{arXiv preprint arXiv:2310.06770}, 2023.
\newblock \doi{10.48550/arXiv.2310.06770}.
\newblock URL \url{https://arxiv.org/abs/2310.06770}.
\newblock Published at ICLR 2024.

\bibitem[Karpathy(2026)]{karpathy2026llmwiki}
Andrej Karpathy.
\newblock Llm wiki.
\newblock GitHub Gist, April 2026.
\newblock URL
  \url{https://gist.github.com/karpathy/442a6bf555914893e9891c11519de94f}.

\bibitem[{OpenAI}(2026)]{gpt54}
{OpenAI}.
\newblock Gpt-5.4 thinking system card.
\newblock \url{https://openai.com/index/gpt-5-4-thinking-system-card/}, March
  2026.

\bibitem[Ouyang et~al.(2025)Ouyang, Yan, Hsu, Chen, Jiang, Wang, Han, Le,
  Daruki, Tang, Tirumalashetty, Lee, Rofouei, Lin, Han, Lee, and
  Pfister]{reasoningbank}
Siru Ouyang, Jun Yan, I-Hung Hsu, Yanfei Chen, Ke~Jiang, Zifeng Wang, Rujun
  Han, Long~T. Le, Samira Daruki, Xiangru Tang, Vishy Tirumalashetty, George
  Lee, Mahsan Rofouei, Hangfei Lin, Jiawei Han, Chen-Yu Lee, and Tomas Pfister.
\newblock Reasoningbank: Scaling agent self-evolving with reasoning memory.
\newblock \emph{arXiv preprint arXiv:2509.25140}, 2025.
\newblock \doi{10.48550/arXiv.2509.25140}.
\newblock URL \url{https://arxiv.org/abs/2509.25140}.

\bibitem[Packer et~al.(2023)Packer, Wooders, Lin, Fang, Patil, Stoica, and
  Gonzalez]{memgpt}
Charles Packer, Sarah Wooders, Kevin Lin, Vivian Fang, Shishir~G. Patil, Ion
  Stoica, and Joseph~E. Gonzalez.
\newblock Memgpt: Towards llms as operating systems.
\newblock \emph{arXiv preprint arXiv:2310.08560}, 2023.
\newblock \doi{10.48550/arXiv.2310.08560}.
\newblock URL \url{https://arxiv.org/abs/2310.08560}.

\bibitem[Shi et~al.(2026)Shi, Li, Ma, Tu, Wu, Xu, Yang, Wu, Xie, Wu, Zeng,
  Heinrich, Zhang, Chen, Li, and Chang]{saasbench}
Kean Shi, Zihang Li, Tianyi Ma, Zengji Tu, Jialong Wu, Xinbo Xu, Qingyao Yang,
  Ruoyu Wu, Weichu Xie, Ming Wu, Jason Zeng, Michael Heinrich, Elvis Zhang,
  Liang Chen, Kuan Li, and Baobao Chang.
\newblock {SaaS-Bench}: Can computer-use agents leverage real-world {SaaS} to
  solve professional workflows?
\newblock \emph{arXiv preprint arXiv:2605.15777}, 2026.
\newblock \doi{10.48550/arXiv.2605.15777}.
\newblock URL \url{https://arxiv.org/abs/2605.15777}.

\bibitem[Shinn et~al.(2023)Shinn, Cassano, Berman, Gopinath, Narasimhan, and
  Yao]{reflexion}
Noah Shinn, Federico Cassano, Edward Berman, Ashwin Gopinath, Karthik
  Narasimhan, and Shunyu Yao.
\newblock Reflexion: Language agents with verbal reinforcement learning.
\newblock \emph{arXiv preprint arXiv:2303.11366}, 2023.
\newblock \doi{10.48550/arXiv.2303.11366}.
\newblock URL \url{https://arxiv.org/abs/2303.11366}.

\bibitem[Squire(1992)]{squire1992}
Larry~R. Squire.
\newblock Declarative and nondeclarative memory: Multiple brain systems
  supporting learning and memory.
\newblock \emph{Journal of Cognitive Neuroscience}, 4\penalty0 (3):\penalty0
  232--243, 1992.

\bibitem[Sumers et~al.(2023)Sumers, Yao, Narasimhan, and Griffiths]{coala}
Theodore~R. Sumers, Shunyu Yao, Karthik Narasimhan, and Thomas~L. Griffiths.
\newblock Cognitive architectures for language agents.
\newblock \emph{arXiv preprint arXiv:2309.02427}, 2023.
\newblock \doi{10.48550/arXiv.2309.02427}.
\newblock URL \url{https://arxiv.org/abs/2309.02427}.

\bibitem[Tulving(1972)]{tulving1972}
Endel Tulving.
\newblock Episodic and semantic memory.
\newblock In Endel Tulving and Wayne Donaldson (eds.), \emph{Organization of
  Memory}, pp.\  381--403. Academic Press, 1972.

\bibitem[Wang et~al.(2026)Wang, Yu, Xie, Yao, Fang, Qiao, Cao, Zheng, Qi,
  Zhang, and Deng]{skillx}
Chenxi Wang, Zhuoyun Yu, Xin Xie, Wuguannan Yao, Runnan Fang, Shuofei Qiao,
  Kexin Cao, Guozhou Zheng, Xiang Qi, Peng Zhang, and Shumin Deng.
\newblock Skillx: Automatically constructing skill knowledge bases for agents.
\newblock \emph{arXiv preprint arXiv:2604.04804}, 2026.
\newblock \doi{10.48550/arXiv.2604.04804}.
\newblock URL \url{https://arxiv.org/abs/2604.04804}.

\bibitem[Wang et~al.(2023)Wang, Xie, Jiang, Mandlekar, Xiao, Zhu, Fan, and
  Anandkumar]{voyager}
Guanzhi Wang, Yuqi Xie, Yunfan Jiang, Ajay Mandlekar, Chaowei Xiao, Yuke Zhu,
  Linxi Fan, and Anima Anandkumar.
\newblock Voyager: An open-ended embodied agent with large language models.
\newblock \emph{arXiv preprint arXiv:2305.16291}, 2023.
\newblock \doi{10.48550/arXiv.2305.16291}.
\newblock URL \url{https://arxiv.org/abs/2305.16291}.

\bibitem[Wang et~al.(2024)Wang, Mao, Fried, and Neubig]{awm}
Zora~Zhiruo Wang, Jiayuan Mao, Daniel Fried, and Graham Neubig.
\newblock Agent workflow memory.
\newblock \emph{arXiv preprint arXiv:2409.07429}, 2024.
\newblock \doi{10.48550/arXiv.2409.07429}.
\newblock URL \url{https://arxiv.org/abs/2409.07429}.

\bibitem[Xu et~al.(2024)Xu, Song, Li, Tang, Jain, Bao, Wang, Zhou, Guo, Cao,
  Yang, Lu, Martin, Su, Maben, Mehta, Chi, Jang, Xie, Zhou, and
  Neubig]{theagentcompany}
Frank~F. Xu, Yufan Song, Boxuan Li, Yuxuan Tang, Kritanjali Jain, Mengxue Bao,
  Zora~Z. Wang, Xuhui Zhou, Zhitong Guo, Murong Cao, Mingyang Yang, Hao~Yang
  Lu, Amaad Martin, Zhe Su, Leander Maben, Raj Mehta, Wayne Chi, Lawrence Jang,
  Yiqing Xie, Shuyan Zhou, and Graham Neubig.
\newblock Theagentcompany: Benchmarking llm agents on consequential real world
  tasks.
\newblock \emph{arXiv preprint arXiv:2412.14161}, 2024.
\newblock \doi{10.48550/arXiv.2412.14161}.
\newblock URL \url{https://arxiv.org/abs/2412.14161}.

\bibitem[Xu et~al.(2025)Xu, Liang, Mei, Gao, Tan, and Zhang]{amem}
Wujiang Xu, Zujie Liang, Kai Mei, Hang Gao, Juntao Tan, and Yongfeng Zhang.
\newblock A-mem: Agentic memory for llm agents.
\newblock \emph{arXiv preprint arXiv:2502.12110}, 2025.
\newblock \doi{10.48550/arXiv.2502.12110}.
\newblock URL \url{https://arxiv.org/abs/2502.12110}.

\bibitem[Zhao et~al.(2023)Zhao, Huang, Xu, Lin, Liu, and Huang]{expel}
Andrew Zhao, Daniel Huang, Quentin Xu, Matthieu Lin, Yong-Jin Liu, and Gao
  Huang.
\newblock Expel: Llm agents are experiential learners.
\newblock \emph{arXiv preprint arXiv:2308.10144}, 2023.
\newblock \doi{10.48550/arXiv.2308.10144}.
\newblock URL \url{https://arxiv.org/abs/2308.10144}.
\newblock Published at AAAI 2024.

\bibitem[Zhou et~al.(2025)Zhou, Chen, Guo, Yan, Lee, Wang, Lee, Zhang, Shao,
  Yang, and Wang]{memento}
Huichi Zhou, Yihang Chen, Siyuan Guo, Xue Yan, Kin~Hei Lee, Zihan Wang, Ka~Yiu
  Lee, Guchun Zhang, Kun Shao, Linyi Yang, and Jun Wang.
\newblock Memento: Fine-tuning llm agents without fine-tuning llms.
\newblock \emph{arXiv preprint arXiv:2508.16153}, 2025.
\newblock \doi{10.48550/arXiv.2508.16153}.
\newblock URL \url{https://arxiv.org/abs/2508.16153}.

\end{thebibliography}
\bibliographystyle{iclr2027_conference}

\newpage
\appendix

\section{Extended Formalization}
\label{app:formalization}

\paragraph{Harness and residual knowledge.}
The fixed harness $H$ contains the executor implementation, system prompt,
available tools, environment interface, and retrieval mechanism. In our
implementation, $H$ is instantiated by \texttt{pi-agent}, and its action space
contains only bash commands. We use \emph{residual procedural knowledge} for
any reusable condition, action, implementation detail, or error path that
improves execution but is absent from $H$. This definition is operational: a
knowledge item is residual if it can be learned from a rollout and changes
later task execution without changing the model or harness.

\paragraph{Round update.}
For the $n=127$ benchmark tasks $Q=\{q_i\}_{i=1}^{n}$, the execute phase under $M_r$
produces rollout batch $B_r$, the set $\{\tau_i^r\}_{i=1}^{n}$ in which
$\tau_i^r$ is the trace of the attempt at $q_i$ during round $r$. A feedback function $E$ maps each rollout and its
resulting environment state to $f_i^r$. Scope $S$ selects passed checkpoint
evidence only or both passed and failed checkpoint evidence. The consolidate
phase applies
\[
M_{r+1}=C\!\left(M_r,
  \{(\tau_i^r,f_i^r)\mid q_i\in Q\};S\right).
\]
The memory store remains frozen within an execute phase, so all rollouts in one
batch use the same $M_r$; updates become visible only in the next round.

\paragraph{Procedural sufficiency.}
A consolidated procedure should specify (i) its purpose, (ii) conditions under
which it applies, (iii) executable actions and implementation details, and
(iv) recovery behavior when an action fails. Consolidation is successful when
it transforms multiple rollout-specific observations into a procedure that
improves execution on other tasks sharing those conditions. This criterion
permits different surface representations while excluding pure replay as the
target behavior.

\begin{table}[th]
\caption{\textbf{Memory types and evaluation requirements.} Procedural memory
differs because repeated trials should refine a reusable action policy, and its
quality is established by execution rather than recall.}
\label{tab:memory-types}
\centering
\small
\begin{tabularx}{\linewidth}{@{}lXXX@{}}
\toprule
 & Episodic & Semantic & Procedural \\
\midrule
Stored content &
Specific event and context &
General fact or concept &
Condition-action procedure \\
Repeated trials &
Remain distinct events &
Duplicate facts merge &
Refine one reusable procedure \\
Failure value &
Another event &
Wrong facts are discarded &
Reveals missing conditions or error paths \\
Generalization target &
Event recall &
Concept-level recall &
Execution across task instantiations \\
Primary evaluation &
Recall accuracy &
Recall accuracy &
Execution success \\
\bottomrule
\end{tabularx}
\end{table}

\section{Memory and Experimental Conditions}
\label{app:methods}

\subsection{Consolidation Methods}

\begin{table}[h]
\caption{\textbf{Consolidation-method definitions.} ``Existing store access''
refers to whether the consolidator reads the current memory store before proposing an
update. \cond{SkillX} and \cond{Insights} are alternative consolidation
representations, whereas \cond{Schema Form-ED}, \cond{Schema Form}, and
\cond{Free Form} are LLM wiki variants.}
\label{tab:method-definitions}
\centering
\resizebox{\linewidth}{!}{%
\begin{tabular}{lllll}
\toprule
Method & Durable representation & Input grouping &
Existing store access & Organization / deduplication \\
\midrule
No Memory & none & none & no & none \\
No Consolidation & raw rollout & rollout & no & raw-rollout retrieval \\
SkillX & planning/functional/atomic skills & rollout independently & no &
three levels, no deduplication \\
Insights & strategy insight & rollout independently & no & flat, no deduplication \\
Schema Form-ED & typed wiki page & independent extraction then merge & merge step only &
fixed schema, deterministic matching \\
Schema Form & typed wiki page & sequential rollouts & yes &
fixed schema, implicit deduplication \\
Free Form & unconstrained wiki page & sequential rollouts & yes &
LLM-managed organization, implicit deduplication \\
\bottomrule
\end{tabular}}
\end{table}

Every row of the matrix in Table~\ref{tab:experiment-matrix} is run for
Claude Opus 4.6, GPT-5.4, and GLM-5 under the same benchmark, executor
loop, and task budget.

\begin{figure}[t]
\centering
\includegraphics[width=0.92\linewidth]{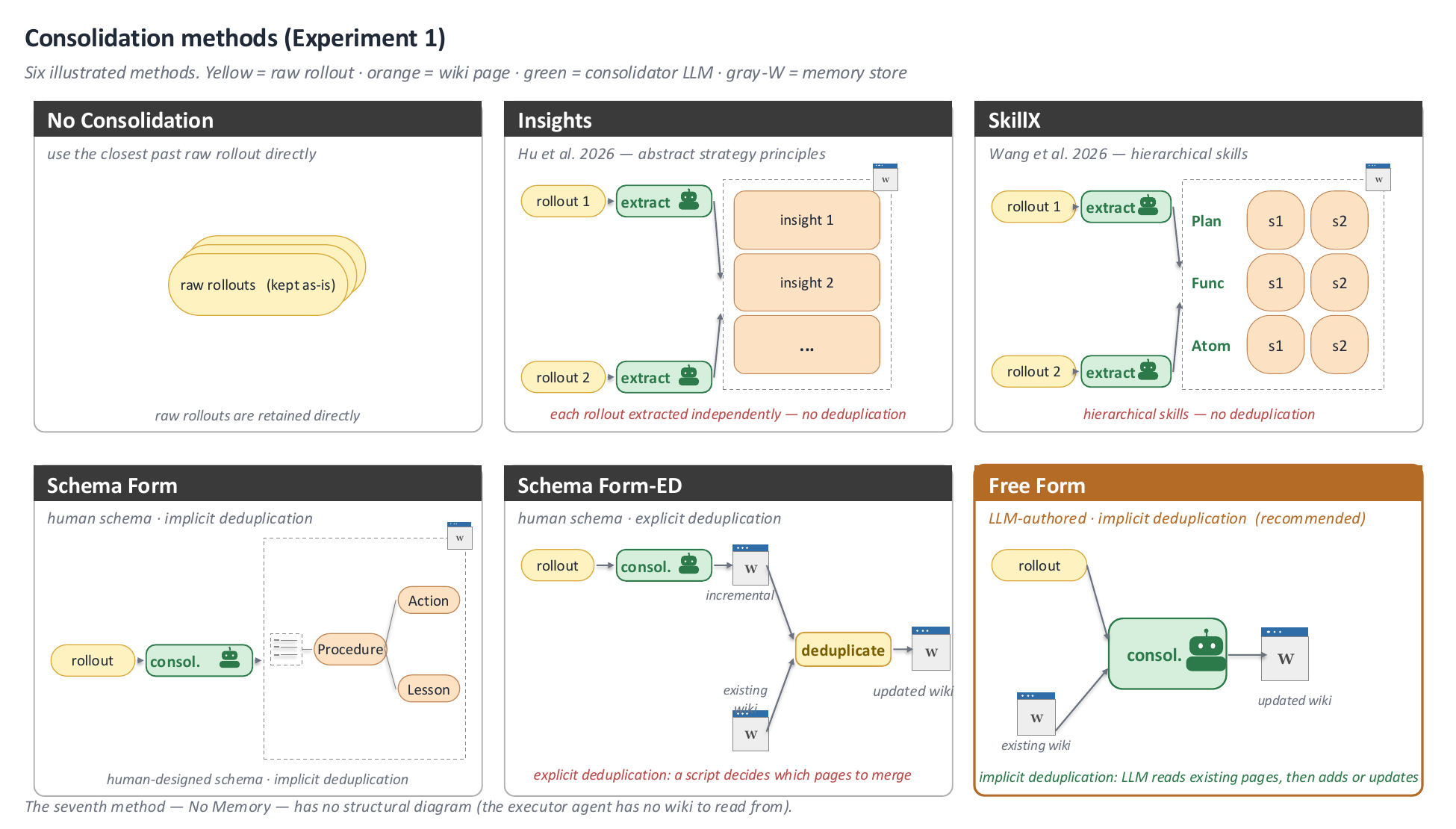}

\vspace{0.6em}
\begin{minipage}[t]{0.64\linewidth}
    \centering
    \includegraphics[width=\linewidth]{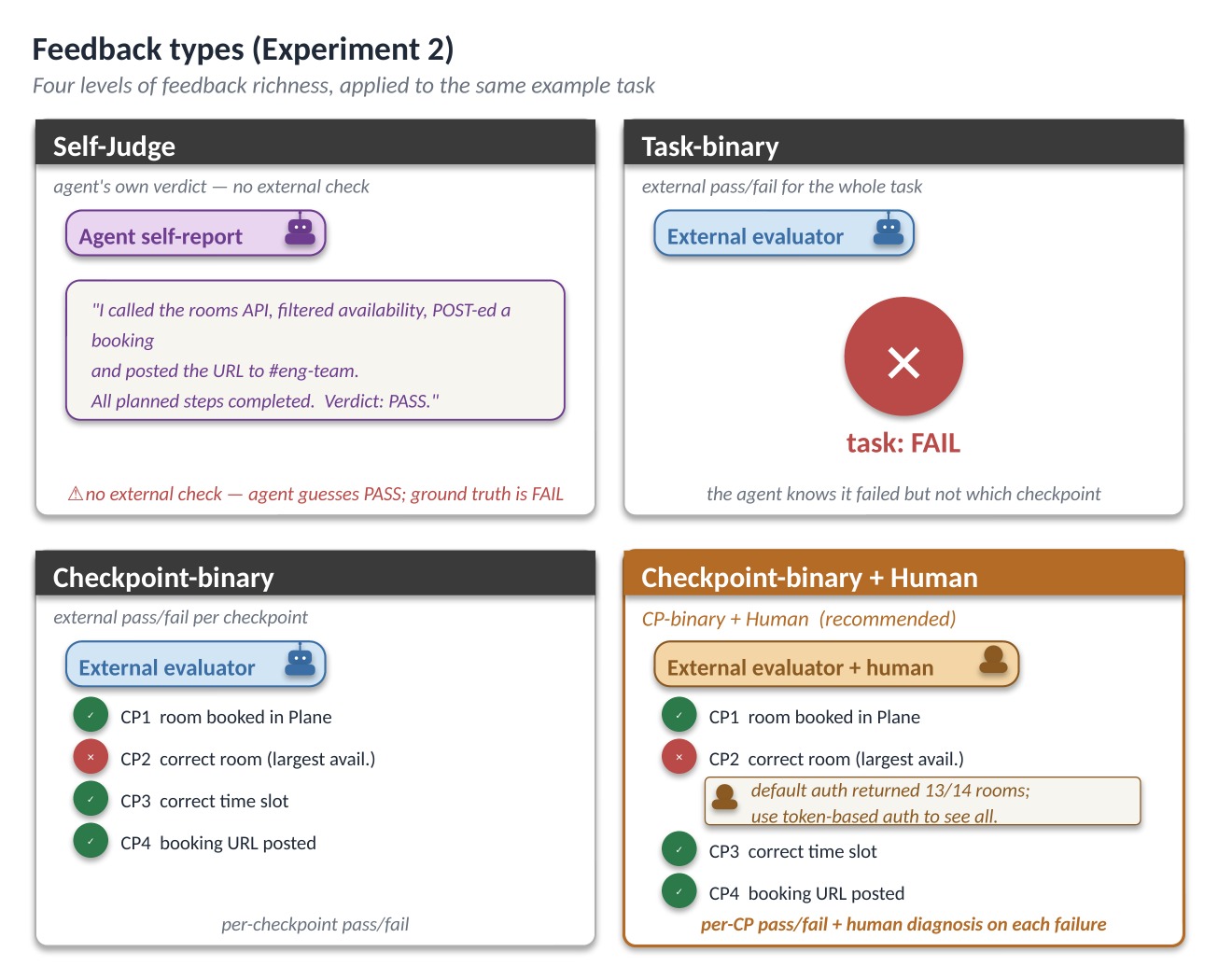}
\end{minipage}\hfill
\begin{minipage}[t]{0.32\linewidth}
    \centering
    \includegraphics[width=\linewidth]{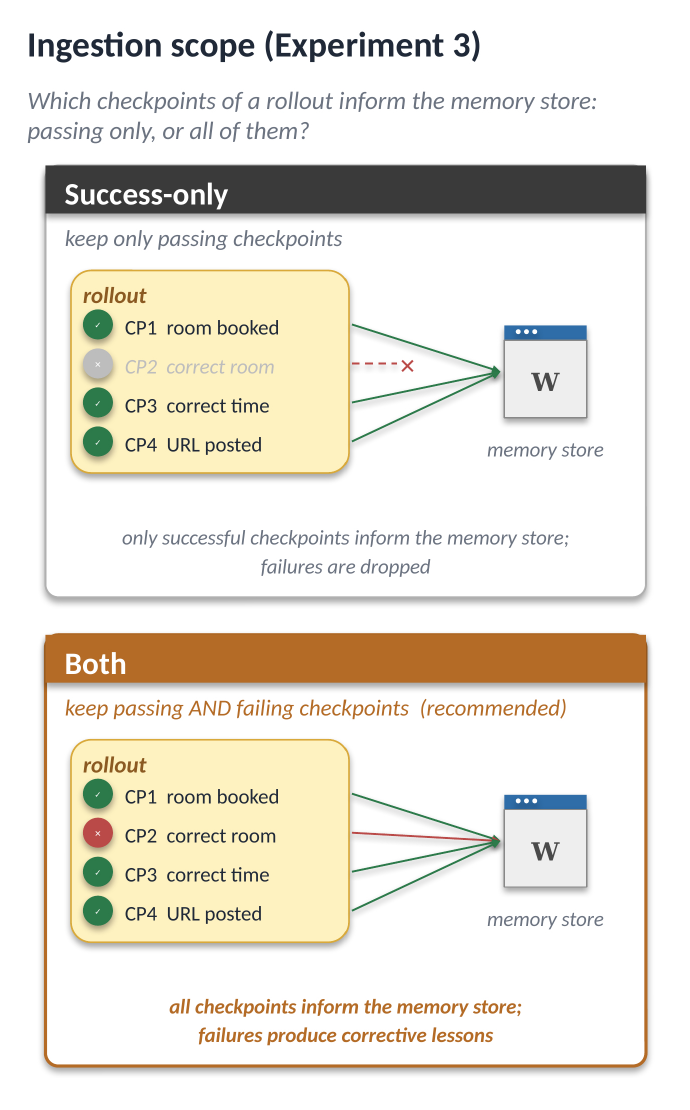}
\end{minipage}
\caption{\textbf{Controlled procedural-memory design space.}
The consolidation methods varied in Experiment~1 appear above. Feedback
type (Experiment~2) and ingestion scope (Experiment~3) appear below at left and
right, respectively. The repeated workflow in Figure~\ref{fig:overview-workflow}
defines the round dimension in Experiment~4.}
\label{fig:overview-design}
\end{figure}

\paragraph{\cond{No Memory} and \cond{No Consolidation}.}
\cond{No Memory} executes without persistent experience. \cond{No Consolidation} stores raw
rollouts and retrieves the most similar one, testing whether experience replay
alone is sufficient.

\paragraph{\cond{Insights} and \cond{SkillX}.}
\cond{Insights} \citep{insights} independently extracts abstract strategy principles
from each rollout and gathers them in a flat store without cross-rollout
deduplication. \cond{SkillX} \citep{skillx} independently extracts planning,
functional, and atomic skills and organizes them by level, also without
cross-rollout deduplication.

\paragraph{LLM wiki variants.}
\cond{Free Form} lets the consolidator inspect the current wiki, then
decide what to record and whether to add a page or revise one. \cond{Schema Form} retains this one-step,
implicit deduplication process but requires three human-defined page types:
Procedures, Actions, and Lessons. \cond{Schema Form-ED}, short for Schema
Form with explicit deduplication, uses the same page schema but
first extracts incremental pages without seeing existing content. A
deterministic script then matches page identifiers and asks the LLM to execute
the selected merges.

\paragraph{Implicit and explicit deduplication.}
The two settings differ in who decides that a new page and an existing
page describe the same thing. Under \emph{implicit} deduplication, used by \cond{Free Form} and \cond{Schema Form}, consolidation is a single stage. The
consolidator sees the current wiki together with the new evidence. Deciding
whether that evidence
extends an existing page or warrants a new one is part of the same judgment
that writes the page. Under \emph{explicit} deduplication, used by \cond{Schema Form-ED}, consolidation is two stages. The consolidator first writes pages for
the new evidence with no sight of the current wiki, producing an incremental
wiki. A deterministic script then normalizes every page identifier and treats
two pages as duplicates when their normalized identifiers agree, for example
mapping any page describing a shell invocation of the Python interpreter to the
same action identifier regardless of what that invocation was for. Only the
selected pairs are passed back to the LLM, whose role is reduced to merging two
pages it is told belong together.

\paragraph{Page linking.}
Every wiki variant is reached from a single \texttt{index.md} entry
point and connected by Obsidian-style \texttt{[[wikilinks]]}, so that retrieval
can start at the index and walk to related content. The variants differ in what
the links are allowed to express. \cond{Free Form} places links wherever the
consolidator judges two pages related, and the resulting graph is unconstrained.
The two schema variants inherit a fixed link grammar from the page types: a
Procedure links to every Action it invokes, and a Lesson attaches to the single
Procedure or Action whose failure it describes. Actions do not link to one
another, so a procedure's steps are recoverable only through the Procedure page
that lists them. Table~\ref{tab:qualitative-filter} records a related
omission, a schema page that names a step without the detail needed to execute
it.

\subsection{Memory Organization by Method}

\paragraph{\cond{No Consolidation}: raw rollout store.}

\paragraph{\cond{Insights}: flat insight set.}

\paragraph{\cond{SkillX}: three skill tiers.}

\paragraph{\cond{Schema Form} and \cond{Schema Form-ED}: typed wiki pages.}

\paragraph{\cond{Free Form}: unconstrained LLM wiki.}

\subsection{Execution Process}

\paragraph{Executor loop and task budget.}

\paragraph{How retrieved memory enters the executor context.}

\paragraph{Termination and what a rollout record contains.}

\subsection{Consolidation and Retrieval Protocols}

\paragraph{Consolidation execution.}
All task executions within round $r$ use the frozen store $M_r$. After
evaluation, the eligible rollouts, meaning those the ingestion scope
admits, are enumerated in lexicographic filename order.
\cond{Free Form} and \cond{Schema Form} process them sequentially against a mutable copy of
$M_r$, so each successful update is visible when processing subsequent
rollouts. \cond{Insights} and \cond{SkillX} process each rollout independently, potentially
in parallel, without access to other extracted outputs. \cond{Schema Form-ED}
independently extracts candidate pages, applies deterministic matching, and
invokes the LLM for selected merges. After all eligible rollouts are processed,
the semantic index over the store is rebuilt to produce $M_{r+1}$.

\paragraph{Retrieval execution.}
Each condition runs its retrieval stage before planning. \cond{No Memory} omits
retrieval, and \cond{No Consolidation} supplies the retrieved
raw-rollout trace. For
consolidated representations, a retrieval-only agent receives the task
instruction and a read-only memory path in a fresh session. It may search the
representation-specific semantic index, inspect the representation's index or
routing file, and read selected files in full, but it cannot interact with the
task environment. If task-level retrieval finds no matched memory, the agent
may search again for each \emph{subgoal}, meaning one step of the plan
the executor has formed. An execution failure also
triggers a fresh subgoal search. Table~\ref{tab:retrieval-adapters} summarizes the
adapter-specific search and expansion rules.

\begin{table}[h]
\caption{\textbf{Representation-specific retrieval adapters.} The executor
loop is shared, while each adapter exposes the native organization of its
memory representation. ``Search'' is semantic search over file contents,
and ``index'' and ``routing'' are the representation's own entry-point file.}
\label{tab:retrieval-adapters}
\centering
\resizebox{\linewidth}{!}{%
\begin{tabular}{llll}
\toprule
Representation & Task-level retrieval & Subgoal retrieval & Context expansion \\
\midrule
No Memory & none & none & none \\
No Consolidation & raw-rollout retrieval & none & retrieved rollout trace \\
Insights & search/index, up to 5 files & search, up to 3 files & full insight files \\
SkillX & search/routing, up to 5 files & search/routing, up to 3 files &
full files across skill tiers \\
Schema variants & top procedure from 3 candidates & top 5 actions per query &
linked procedures, actions, and lessons \\
Free Form & search top 5 per query/index & search/index, up to 3 pages &
full wiki pages \\
\bottomrule
\end{tabular}}
\end{table}

\paragraph{Retrieval walkthrough for each method.}

\subsection{Feedback Type, Ingestion Scope, and Rounds}

\begin{table}[h]
\caption{\textbf{Feedback types.} Rich feedback denotes \cond{Checkpoint-binary + Human}, and weak feedback denotes \cond{Self-Judge}.}
\label{tab:feedback-definitions}
\centering
\small
\begin{tabularx}{\linewidth}{@{}lX@{}}
\toprule
Feedback type & Information available during consolidation \\
\midrule
Self-Judge & Agent self-assessment without an external outcome signal. \\
Task-binary & External pass/fail for the whole task. \\
Checkpoint-binary & External pass/fail for each checkpoint. \\
Checkpoint-binary + Human & Checkpoint outcomes plus human failure diagnosis. \\
\bottomrule
\end{tabularx}
\end{table}

\cond{Success-only} selects evidence associated with passed checkpoints. \cond{Both}
(successes and failures) selects evidence from all checkpoints. One round is one execute phase over all
benchmark tasks followed by one consolidate phase. R0 has no memory
store, while R1 to R4 execute with and then update the persistent store
produced previously.

\subsection{Controlled Experimental Matrix}

\begin{table}[h]
\caption{\textbf{Controlled experimental matrix.} Each experiment varies one
focal dimension while fixing the remaining dimensions.}
\label{tab:experiment-matrix}
\centering
\resizebox{\linewidth}{!}{%
\begin{tabular}{llllll}
\toprule
Experiment & Varied dimension & Method & Feedback & Scope & Rounds \\
\midrule
1 & Consolidation method & seven conditions &
Checkpoint-binary & Both & R1 \\
2 & Feedback type & Free Form &
four conditions & Both & R1 \\
3 & Ingestion scope & Free Form &
Self-Judge and Checkpoint-binary + Human & Success-only, Both & R1 \\
4 & Rounds & Free Form &
Self-Judge and Checkpoint-binary + Human & Both & R0 to R4 \\
\bottomrule
\end{tabular}}
\end{table}

\subsection{Consolidation Prompts}

\section{Complete Cross-Model Results}
\label{app:results}

Tables~\ref{tab:exp1-full} to~\ref{tab:pooled-full} report the exact result
variants underlying the main-text ranges and figures. Within each model
block, bold marks the best value in each column, where best means the lowest
value for store size and cost. Shading marks the strongest configuration in the
block. Table~\ref{tab:exp4-weak} has none, because there no single round is
strongest on every model.

\subsection{Experiment 1: Consolidation Method}

\begin{table}[p]
\caption{\textbf{Complete Experiment 1 results.} $\Delta$ is checkpoint-pass
improvement over \cond{No Memory} for the same model. Cost covers consolidation API
calls only.}
\label{tab:exp1-full}
\centering
\resizebox{\linewidth}{!}{%
\begin{tabular}{llccccc}
\toprule
Model & Method & Checkpoint pass & $\Delta$ (pp) &
Fully passed & Store (KB) & Consol.\ cost \\
\midrule
Claude Opus 4.6 & No Memory & 317/542 (58.5\%) & n/a & 46/127 (36.2\%) & n/a & n/a \\
& No Consolidation & 332/542 (61.3\%) & +2.8 & 51/127 (40.2\%) & 7{,}979 & n/a \\
& SkillX & 354/542 (65.3\%) & +6.8 & 58/127 (45.7\%) & 1{,}125 & \$173 \\
& Insights & 356/542 (65.7\%) & +7.2 & 58/127 (45.7\%) & 396 & \textbf{\$148} \\
& Schema Form-ED & 356/542 (65.7\%) & +7.2 & 57/127 (44.9\%) & 789 & \$380 \\
& Schema Form & 359/542 (66.2\%) & +7.7 & \textbf{62/127 (48.8\%)} & 565 & \$230 \\
\rowcolor{OurRow}
& \textbf{Free Form} & \textbf{392/542 (72.3\%)} &
\textbf{+13.8} & 61/127 (48.0\%) & \textbf{234} & \$198 \\
\midrule
GPT-5.4 & No Memory & 325/542 (60.0\%) & n/a & 51/127 (40.2\%) & n/a & n/a \\
& No Consolidation & 339/542 (62.5\%) & +2.5 & 54/127 (42.5\%) & 8{,}208 & n/a \\
& SkillX & 364/542 (67.2\%) & +7.2 & 64/127 (50.4\%) & 1{,}161 & \$171 \\
& Insights & 359/542 (66.2\%) & +6.2 & 62/127 (48.8\%) & 399 & \textbf{\$146} \\
& Schema Form-ED & 363/542 (67.0\%) & +7.0 & 63/127 (49.6\%) & 808 & \$370 \\
& Schema Form & 368/542 (67.9\%) & +7.9 & 64/127 (50.4\%) & 592 & \$229 \\
\rowcolor{OurRow}
& \textbf{Free Form} & \textbf{393/542 (72.5\%)} &
\textbf{+12.5} & \textbf{66/127 (52.0\%)} & \textbf{233} & \$202 \\
\midrule
GLM-5 & No Memory & 319/542 (58.9\%) & n/a & 49/127 (38.6\%) & n/a & n/a \\
& No Consolidation & 335/542 (61.8\%) & +2.9 & 52/127 (40.9\%) & 8{,}480 & n/a \\
& SkillX & 356/542 (65.7\%) & +6.8 & 59/127 (46.5\%) & 1{,}187 & \textbf{\$79} \\
& Insights & 357/542 (65.9\%) & +7.0 & 60/127 (47.2\%) & 441 & \$87 \\
& Schema Form-ED & 356/542 (65.7\%) & +6.8 & 59/127 (46.5\%) & 838 & \$197 \\
& Schema Form & 361/542 (66.6\%) & +7.7 & 64/127 (50.4\%) & 599 & \$117 \\
\rowcolor{OurRow}
& \textbf{Free Form} & \textbf{393/542 (72.5\%)} &
\textbf{+13.6} & \textbf{66/127 (52.0\%)} & \textbf{255} & \$118 \\
\bottomrule
\end{tabular}}
\end{table}

\subsection{Experiment 2: Feedback Type}

\begin{table}[p]
\caption{\textbf{Complete Experiment 2 results.} $\Delta$ is relative to
\cond{Self-Judge} for the same model.}
\label{tab:exp2-full}
\centering
\resizebox{\linewidth}{!}{%
\begin{tabular}{llccccc}
\toprule
Model & Feedback & Checkpoint pass & $\Delta$ (pp) &
Fully passed & Store (KB) & Consol.\ cost \\
\midrule
Claude Opus 4.6 & Self-Judge & 352/542 (64.9\%) & 0.0 & 59/127 (46.5\%) & 212 & \$174 \\
& Task-binary & 391/542 (72.1\%) & +7.2 & 64/127 (50.4\%) & 237 & \$205 \\
& Checkpoint-binary & 392/542 (72.3\%) & +7.4 & 61/127 (48.0\%) & 234 & \$198 \\
\rowcolor{OurRow}
& \textbf{Checkpoint-binary + Human} &
\textbf{411/542 (75.8\%)} & \textbf{+10.9} &
\textbf{71/127 (55.9\%)} & 328 & \$196 \\
\midrule
GPT-5.4 & Self-Judge & 352/542 (64.9\%) & 0.0 & 56/127 (44.1\%) & 200 & \$156 \\
& Task-binary & 384/542 (70.8\%) & +5.9 & 66/127 (52.0\%) & 224 & \$186 \\
& Checkpoint-binary & 393/542 (72.5\%) & +7.6 & 66/127 (52.0\%) & 233 & \$202 \\
\rowcolor{OurRow}
& \textbf{Checkpoint-binary + Human} &
\textbf{418/542 (77.1\%)} & \textbf{+12.2} &
\textbf{78/127 (61.4\%)} & 335 & \$201 \\
\midrule
GLM-5 & Self-Judge & 349/542 (64.4\%) & 0.0 & 56/127 (44.1\%) & 235 & \$97 \\
& Task-binary & 388/542 (71.6\%) & +7.2 & 64/127 (50.4\%) & 256 & \$106 \\
& Checkpoint-binary & 393/542 (72.5\%) & +8.1 & 66/127 (52.0\%) & 255 & \$118 \\
\rowcolor{OurRow}
& \textbf{Checkpoint-binary + Human} &
\textbf{414/542 (76.4\%)} & \textbf{+12.0} &
\textbf{72/127 (56.7\%)} & 343 & \$103 \\
\bottomrule
\end{tabular}}
\end{table}

\subsection{Experiment 3: Ingestion Scope}

\begin{table}[p]
\caption{\textbf{Complete Experiment 3 results.} $\Delta$ is relative to
\cond{Success-only} under the same model and feedback type.
CP-binary + Human abbreviates \cond{Checkpoint-binary + Human}.}
\label{tab:exp3-full}
\centering
\resizebox{\linewidth}{!}{%
\begin{tabular}{lllccccc}
\toprule
Model & Feedback & Scope & Checkpoint pass & $\Delta$ (pp) &
Fully passed & Store (KB) & Consol.\ cost \\
\midrule
Claude Opus 4.6 & CP-binary + Human & Success-only &
359/542 (66.2\%) & 0.0 & 53/127 (41.7\%) & 119 & \$180 \\
\rowcolor{OurRow}
& \textbf{CP-binary + Human} & \textbf{Both} &
\textbf{411/542 (75.8\%)} & \textbf{+9.6} &
\textbf{71/127 (55.9\%)} & 328 & \$196 \\
& Self-Judge & Success-only &
351/542 (64.8\%) & 0.0 & 54/127 (42.5\%) & 193 & \$180 \\
& Self-Judge & Both &
352/542 (64.9\%) & +0.1 & 59/127 (46.5\%) & 212 & \$174 \\
\midrule
GPT-5.4 & CP-binary + Human & Success-only &
361/542 (66.6\%) & 0.0 & 60/127 (47.2\%) & 135 & \$170 \\
\rowcolor{OurRow}
& \textbf{CP-binary + Human} & \textbf{Both} &
\textbf{418/542 (77.1\%)} & \textbf{+10.5} &
\textbf{78/127 (61.4\%)} & 335 & \$201 \\
& Self-Judge & Success-only &
356/542 (65.7\%) & 0.0 & 58/127 (45.7\%) & 211 & \$173 \\
& Self-Judge & Both &
352/542 (64.9\%) & $-$0.8 & 56/127 (44.1\%) & 200 & \$156 \\
\midrule
GLM-5 & CP-binary + Human & Success-only &
362/542 (66.8\%) & 0.0 & 55/127 (43.3\%) & 140 & \$100 \\
\rowcolor{OurRow}
& \textbf{CP-binary + Human} & \textbf{Both} &
\textbf{414/542 (76.4\%)} & \textbf{+9.6} &
\textbf{72/127 (56.7\%)} & 343 & \$103 \\
& Self-Judge & Success-only &
351/542 (64.8\%) & 0.0 & 56/127 (44.1\%) & 212 & \$91 \\
& Self-Judge & Both &
349/542 (64.4\%) & -0.4 & 56/127 (44.1\%) & 235 & \$97 \\
\bottomrule
\end{tabular}}
\end{table}

\subsection{Experiment 4: Rounds}

\begin{table}[p]
\caption{\textbf{Experiment 4 with rich feedback.} $\Delta$ is relative to
R0. Marginal gain is relative to the preceding round.}
\label{tab:exp4-rich}
\centering
\resizebox{\linewidth}{!}{%
\begin{tabular}{llcccccc}
\toprule
Model & Round & Checkpoint pass & $\Delta$ (pp) & Marginal (pp) &
Fully passed & Store (KB) & Consol.\ cost \\
\midrule
Claude Opus 4.6 & R0 & 317/542 (58.5\%) & 0.0 & n/a & 46/127 (36.2\%) & n/a & n/a \\
& R1 & 411/542 (75.8\%) & +17.3 & +17.3 & 71/127 (55.9\%) & 328 & \$196 \\
& R2 & 430/542 (79.3\%) & +20.8 & +3.5 & 79/127 (62.2\%) & 460 & \$171 \\
& R3 & 446/542 (82.3\%) & +23.8 & +3.0 & 86/127 (67.7\%) & 492 & \$180 \\
\rowcolor{OurRow}
& \textbf{R4} & \textbf{464/542 (85.6\%)} &
\textbf{+27.1} & +3.3 & \textbf{92/127 (72.4\%)} & 583 & \$179 \\
\midrule
GPT-5.4 & R0 & 325/542 (60.0\%) & 0.0 & n/a & 51/127 (40.2\%) & n/a & n/a \\
& R1 & 418/542 (77.1\%) & +17.1 & +17.1 & 78/127 (61.4\%) & 335 & \$201 \\
& R2 & 438/542 (80.8\%) & +20.8 & +3.7 & 86/127 (67.7\%) & 465 & \$165 \\
& R3 & 449/542 (82.8\%) & +22.8 & +2.0 & 91/127 (71.7\%) & 505 & \$178 \\
\rowcolor{OurRow}
& \textbf{R4} & \textbf{470/542 (86.7\%)} &
\textbf{+26.7} & +3.9 & \textbf{99/127 (78.0\%)} & 575 & \$170 \\
\midrule
GLM-5 & R0 & 319/542 (58.9\%) & 0.0 & n/a & 49/127 (38.6\%) & n/a & n/a \\
& R1 & 414/542 (76.4\%) & +17.5 & +17.5 & 72/127 (56.7\%) & 343 & \$103 \\
& R2 & 433/542 (79.9\%) & +21.0 & +3.5 & 82/127 (64.6\%) & 486 & \$92 \\
& R3 & 451/542 (83.2\%) & +24.3 & +3.3 & 87/127 (68.5\%) & 499 & \$104 \\
\rowcolor{OurRow}
& \textbf{R4} & \textbf{468/542 (86.3\%)} &
\textbf{+27.4} & +3.1 & \textbf{94/127 (74.0\%)} & 603 & \$96 \\
\bottomrule
\end{tabular}}
\end{table}

\begin{table}[p]
\caption{\textbf{Experiment 4 with weak feedback.} $\Delta$ is relative to R0
for the same model.}
\label{tab:exp4-weak}
\centering
\resizebox{\linewidth}{!}{%
\begin{tabular}{llccccc}
\toprule
Model & Round & Checkpoint pass & $\Delta$ (pp) &
Fully passed & Store (KB) & Consol.\ cost \\
\midrule
Claude Opus 4.6 & R0 & 317/542 (58.5\%) & 0.0 & 46/127 (36.2\%) & n/a & n/a \\
& R1 & 352/542 (64.9\%) & +6.4 & 59/127 (46.5\%) & 212 & \$174 \\
& R2 & 349/542 (64.4\%) & +5.9 & 52/127 (40.9\%) & 297 & \$141 \\
& R3 & 362/542 (66.8\%) & +8.3 & 56/127 (44.1\%) & 273 & \$136 \\
& \textbf{R4} & \textbf{373/542 (68.8\%)} & \textbf{+10.3} &
\textbf{60/127 (47.2\%)} & 291 & \$135 \\
\midrule
GPT-5.4 & R0 & 325/542 (60.0\%) & 0.0 & 51/127 (40.2\%) & n/a & n/a \\
& R1 & 352/542 (64.9\%) & +4.9 & 56/127 (44.1\%) & 200 & \$156 \\
& R2 & 358/542 (66.1\%) & +6.1 & 58/127 (45.7\%) & 262 & \$133 \\
& \textbf{R3} & \textbf{364/542 (67.2\%)} & \textbf{+7.2} & \textbf{62/127 (48.8\%)} & 267 & \$133 \\
& R4 & 363/542 (67.0\%) & +7.0 &
62/127 (48.8\%) & 285 & \$144 \\
\midrule
GLM-5 & R0 & 319/542 (58.9\%) & 0.0 & 49/127 (38.6\%) & n/a & n/a \\
& R1 & 349/542 (64.4\%) & +5.5 & 56/127 (44.1\%) & 235 & \$97 \\
& R2 & 347/542 (64.0\%) & +5.1 & 55/127 (43.3\%) & 255 & \$62 \\
& R3 & 345/542 (63.7\%) & +4.8 & 55/127 (43.3\%) & 262 & \$71 \\
& \textbf{R4} & \textbf{355/542 (65.5\%)} & \textbf{+6.6} &
\textbf{57/127 (44.9\%)} & 282 & \$97 \\
\bottomrule
\end{tabular}}
\end{table}

\begin{table}[p]
\caption{\textbf{Rollout-volume control.} The pooled methods consume 635
rollouts from rounds R0 to R4, more than the 508 the evaluated \cond{Free Form}
R4 store was built from, and process them in a single non-incremental pass while
\cond{Free Form} consolidates one batch at a time. $\Delta$ is relative to
\cond{No Memory} for the same model in Table~\ref{tab:exp1-full}.}
\label{tab:pooled-full}
\centering
\resizebox{\linewidth}{!}{%
\begin{tabular}{llccc}
\toprule
Model & Condition & Checkpoint pass & $\Delta$ (pp) & Fully passed \\
\midrule
Claude Opus 4.6 & Pooled R0 to R4, No Consolidation &
408/542 (75.3\%) & +16.8 & 70/127 (55.1\%) \\
& Pooled R0 to R4, Insights & 431/542 (79.5\%) & +21.0 & 78/127 (61.4\%) \\
& Free Form, R2 & 430/542 (79.3\%) & +20.8 & 79/127 (62.2\%) \\
\rowcolor{OurRow}
& \textbf{Free Form, R4} & \textbf{464/542 (85.6\%)} &
\textbf{+27.1} & \textbf{92/127 (72.4\%)} \\
\midrule
GPT-5.4 & Pooled R0 to R4, No Consolidation &
412/542 (76.0\%) & +16.0 & 77/127 (60.6\%) \\
& Pooled R0 to R4, Insights & 433/542 (79.9\%) & +19.9 & 84/127 (66.1\%) \\
& Free Form, R2 & 438/542 (80.8\%) & +20.8 & 86/127 (67.7\%) \\
\rowcolor{OurRow}
& \textbf{Free Form, R4} & \textbf{470/542 (86.7\%)} &
\textbf{+26.7} & \textbf{99/127 (78.0\%)} \\
\midrule
GLM-5 & Pooled R0 to R4, No Consolidation &
406/542 (74.9\%) & +16.0 & 74/127 (58.3\%) \\
& Pooled R0 to R4, Insights & 429/542 (79.2\%) & +20.3 & 81/127 (63.8\%) \\
& Free Form, R2 & 433/542 (79.9\%) & +21.0 & 82/127 (64.6\%) \\
\rowcolor{OurRow}
& \textbf{Free Form, R4} & \textbf{468/542 (86.3\%)} &
\textbf{+27.4} & \textbf{94/127 (74.0\%)} \\
\bottomrule
\end{tabular}}
\end{table}

\section{Additional Analysis}
\label{app:analysis}

\subsection{Store Size and Filter Characterization}

Across methods a larger store does not buy a higher pass rate
(Section~\ref{sec:exp1}). Within a fixed method the same growth can instead
represent useful signal. Across models, richer feedback grows the store from
200 to 235 KB with \cond{Self-Judge} to 328 to 343 KB with
\cond{Checkpoint-binary + Human} while improving checkpoint pass rate
(Table~\ref{tab:exp2-full}). Under rich feedback, repeated rounds grow the
store from 328 to 343 KB at R1 to 575 to 603 KB at R4 while
performance continues to improve (Table~\ref{tab:exp4-rich}). These within-method results distinguish the Filter Hypothesis from
a simple compression objective.

\subsection{Rollout-Volume Control}

\begin{figure}[h]
\centering
\includegraphics[width=0.78\linewidth]{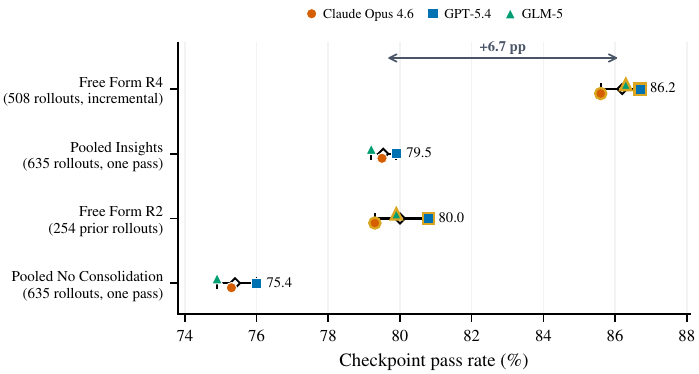}
\caption{\textbf{Rollout volume does not account for the R4 result.}
Colored markers show individual model variants. Black diamonds and
whiskers show the cross-model mean and min to max range.
The pooled controls consume 635 rollouts from R0 to R4, more than the
508 the \cond{Free Form} R4 store was built from, and consolidate them in one
pass while \cond{Free Form} consolidates one batch at a time.
Exact results appear in Table~\ref{tab:pooled-full}.}
\label{fig:rollout-volume-control}
\end{figure}

This control asks whether the R4 result follows from the number of
rollouts consolidated. Pooled \cond{Insights} and \cond{No Consolidation} each
receive 635 rollouts from R0 to R4, more than the 508 the evaluated
\cond{Free Form} R4 store was built from, and consolidate them in a single pass
rather than one batch at a time. On strictly more rollouts, pooled
\cond{Insights} only reaches about what \cond{Free Form} reached from the 254
rollouts behind its R2 store, and it remains 6.1 to 7.1 pp below
\cond{Free Form} R4 for every model
(Figure~\ref{fig:rollout-volume-control}). Rollout volume therefore does not
account for the R4 result. Because the pooled runs also change the consolidation
method, this control rules volume out as the explanation without establishing
incremental consolidation as the cause.

\subsection{Task-Score Distributions}

\begin{figure}[h]
\centering
\includegraphics[width=0.88\linewidth]{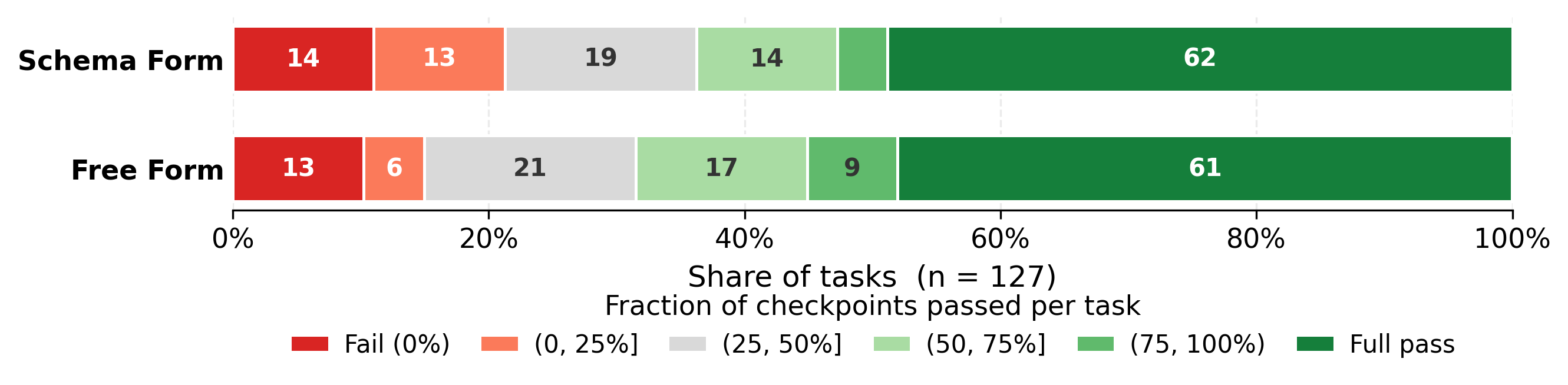}
\caption{\textbf{\cond{Free Form} versus \cond{Schema Form} by task-score bucket for Claude
Opus 4.6.} The checkpoint-level advantage of \cond{Free Form} is distributed across
partially completed tasks, while \cond{Schema Form} narrowly leads in the
all-or-nothing fully-passed rate. This difference motivates reporting both
metrics.}
\label{fig:schema-free-distribution}
\end{figure}

\subsection{Failure-Cause Statistics and Analysis}

\subsection{Task-Score Bucket Counts}

\subsection{Selection of the Qualitative Examples}

\subsection{Qualitative Consolidation Behaviors}

\begin{table}[h]
\caption{\textbf{Representative filtering behaviors observed in generated
stores.} These examples motivate, rather than independently prove, the Filter
Hypothesis.}
\label{tab:qualitative-filter}
\centering
\small
\begin{tabularx}{\linewidth}{@{}lXX@{}}
\toprule
Behavior & Observed pattern & Procedural consequence \\
\midrule
Implicit merge &
Free Form revises one GitLab authentication wiki page with new pagination and error
handling evidence. &
Repeated trials become one richer procedure. \\
Independent duplication &
SkillX creates 15 separate pages describing substantially the same GitLab API
authentication procedure. &
Retrieval must sift through redundant fragments. \\
Schema omission &
On \texttt{ds-stock-analysis-slides}, Schema Form records step names but drops
the code and formatting details retained by Free Form. &
The action is named but not reliably executable. \\
Explicit false merge &
Schema Form-ED merges ``grade ML exams with Python'' and ``extract video frames
with Python'' under \texttt{action-bash-run-python}. &
Unrelated instructions can be retrieved together. \\
\bottomrule
\end{tabularx}
\end{table}

\section{Implementation and Reproducibility}
\label{app:reproducibility}

\paragraph{Environment.}
Each task ran in a Docker environment connected to self-hosted GitLab,
RocketChat, OwnCloud, and Plane services. We used \texttt{pi-agent} v0.75.3 as
the fixed executor harness and exposed bash as its only action
tool. All service interaction, including API calls, was therefore
issued through shell commands. Every attempt was launched with
\texttt{--no-session}, so the harness wrote no session state to disk and
nothing it might otherwise retain could carry across tasks or rounds.
Self-hosted services were confirmed running before each execution batch.

\paragraph{Environment reset between attempts.}

\paragraph{Models and retrieval.}
Claude Opus 4.6, GPT-5.4, and GLM-5 are the executor-consolidator model
variants. Claude Sonnet 4.6 performs evaluation, and Cohere Embed v4
provides 1024-dimensional semantic-search embeddings. The LLM wiki was available through link navigation from
\texttt{index.md} and semantic search over page contents. All conditions used
the same executor loop and a retrieval stage before planning. Consolidated
representations used retrieval-only agents with representation-specific
adapters, while \cond{No Memory} omitted retrieval and \cond{No Consolidation} supplied raw
rollout evidence. Table~\ref{tab:retrieval-adapters} gives the exact
protocols.

\paragraph{Evaluator protocol for checkpoint scoring.}

\paragraph{Compute, time, and human effort.}
Experiments were orchestrated on an Amazon EC2 \texttt{g5.2xlarge} host with
8 vCPUs and 30 GiB of memory. Model inference was served through
provider APIs,
while the host ran the agent harness, benchmark services, and Docker
environments. One complete 127-task evaluation batch occupied approximately 12
wall-clock hours. Producing \cond{Checkpoint-binary + Human} feedback for one rollout
batch required approximately six hours from a qualified annotator.

\begin{table}[t]
\caption{\textbf{API cost of one complete 127-task evaluation batch.}
Execution is the task-execution phase, retrieval is the memory-query phase, and
\cond{Free Form} total is their sum. Consolidation costs are reported separately in
the experiment tables. Evaluator, embedding, host-compute, and
human-feedback costs are outside these values.}
\label{tab:execution-cost}
\centering
\small
\resizebox{\linewidth}{!}{%
\begin{tabular}{lrrrr}
\toprule
Model & No Memory execution & Free Form execution & Retrieval &
Free Form total \\
\midrule
Claude Opus 4.6 & \$106.6 & \$88.9 & \$60.4 & \$149.3 \\
GPT-5.4 & \$108.5 & \$87.1 & \$60.6 & \$147.8 \\
GLM-5 & \$56.3 & \$47.2 & \$18.6 & \$65.9 \\
\bottomrule
\end{tabular}}
\end{table}

\paragraph{Round procedure.}
Within each round, all benchmark tasks executed against a frozen memory store and
produced one rollout batch. Feedback records were generated from checkpoint
evaluation and, for the rich condition, human failure diagnosis. The
consolidator then updated a copy of the previous wiki. The semantic index was
rebuilt before the next execute phase. R0 used no store and supplied the rollout
batch from which R1 memory was created.

\paragraph{Validation.}
Metric denominators were normalized to the benchmark totals for every
condition. Shared conditions reuse one canonical result wherever they appear
across experiments. All main and appendix tables, and the values behind
every figure, are transcribed from the same source-of-truth result pages.

\paragraph{Experiment code.}

\section{Limitations and Future Work}
\label{app:limitations}

Section~\ref{sec:conclusion} lists the bounds on this study. Three of
them constrain the conclusions most, and each is taken up here with the
experiment that would lift it.

\paragraph{Single runs.}
Agent execution is stochastic, but each configuration was executed once. Large
gaps motivate the paper's conclusions. Small differences should be
treated as unresolved until replicated, among them the 0.2 to 1.7 pp between
\cond{Task-binary} and \cond{Checkpoint-binary} (Table~\ref{tab:exp2-full}) and the 0.8 to 1.6 pp between the best
fully-passed rates in Experiment~1 (Table~\ref{tab:exp1-full}). Stateful long-horizon execution makes exhaustive replication
costly, but cost does not remove the need to replicate the central contrasts.

\paragraph{Repeated-task distribution.}
R1 to R4 re-executed the same benchmark tasks. The experiment measures
deepening competence over recurring procedural workloads, not unrestricted
generalization to unseen distributions. Held-out task families and cross-domain
transfer are necessary to measure how broadly the learned procedures apply.

\paragraph{Human diagnosis.}
\cond{Checkpoint-binary + Human} provides the strongest feedback but requires human
effort after execution. Automating failure diagnosis without reverting to
unreliable self-judgment is the most direct route to scaling the approach.

\end{document}